\RequirePackage[svgnames]{xcolor}
\PassOptionsToPackage{table,svgnames}{xcolor}
\PassOptionsToPackage{explicit,compact}{titlesec}
\documentclass[12pt,letterpaper]{mystyle}

\usepackage[all]{hypcap}
\usepackage[svgnames]{xcolor}
\usepackage[comma,authoryear,compress]{natbib}
\usepackage{hyperref}[citecolor=lightblue]

\hypersetup{
    colorlinks = true,
    citecolor = {YaleBlue},
}

\usepackage{algorithm}
\usepackage{algorithmicx}
\usepackage{algpseudocode}
\usepackage{microtype}
\usepackage{graphicx}
\expandafter\def\csname ver@subfig.sty\endcsname{}
\usepackage{booktabs} %
\usepackage{float}
\usepackage{bigstrut}

\usepackage{amsmath}
\usepackage{amssymb}
\usepackage{mathtools}
\usepackage{amsthm}
\usepackage{mathrsfs}
\usepackage{nicefrac}
\usepackage{dsfont}
\usepackage{enumitem}
\usepackage{subcaption}
\usepackage{graphicx,subfig}
\usepackage{cleveref}
\usepackage{bxcoloremoji}

\usepackage{float}

\usepackage[utf8]{inputenc} %
\usepackage[T1]{fontenc}    %
\usepackage{hyperref}       %
\usepackage{url}            %
\usepackage{booktabs}       %
\usepackage{amsfonts}       %
\usepackage{nicefrac}       %
\usepackage{microtype}      %
\usepackage{graphicx}
\usepackage{subcaption} 
\usepackage{amssymb}
\usepackage{fdsymbol}
\usepackage{wrapfig}
\usepackage{lipsum}
\usepackage{enumitem}
\usepackage{stackengine}
\usepackage[font=small,labelfont=bf]{caption}
\usepackage{color}
\usepackage{adjustbox}

\usepackage{rotating}
\usepackage{makecell}

\usepackage{svg}
\usepackage{booktabs}
\usepackage{multirow}

\usepackage{xcolor}   
\usepackage{amssymb}  
\usepackage{ulem}

\newcommand{\x}{\mathbf{x}}

\definecolor{blanchedalmond}{rgb}{1.0, 0.92, 0.8}
\definecolor{carmine}{rgb}{0.59, 0.0, 0.09}
\definecolor{lightblue}{rgb}{0.22,0.45,0.70}%

\renewcommand{\mathbf}{\boldsymbol}

\makeatletter
\def\Ddots{\mathinner{\mkern1mu\raise\p@
\vbox{\kern7\p@\hbox{.}}\mkern2mu
\raise4\p@\hbox{.}\mkern2mu\raise7\p@\hbox{.}\mkern1mu}}
\makeatother

\definecolor{amaranth}{rgb}{0.9, 0.17, 0.31}
\definecolor{antiquebrass}{rgb}{0.8, 0.58, 0.46}
\definecolor{antiquefuchsia}{rgb}{0.57, 0.36, 0.51}
\definecolor{chromeyellow}{rgb}{0.31, 0.47, 0.26}

\newcommand{\github}{\raisebox{-1.5pt}{\includegraphics[height=1.05em]{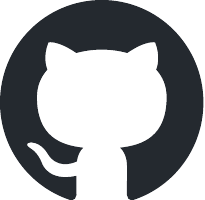}}}

\newcommand{\paperlogo}{\raisebox{-1.5pt}{\includegraphics[height=2.05em]{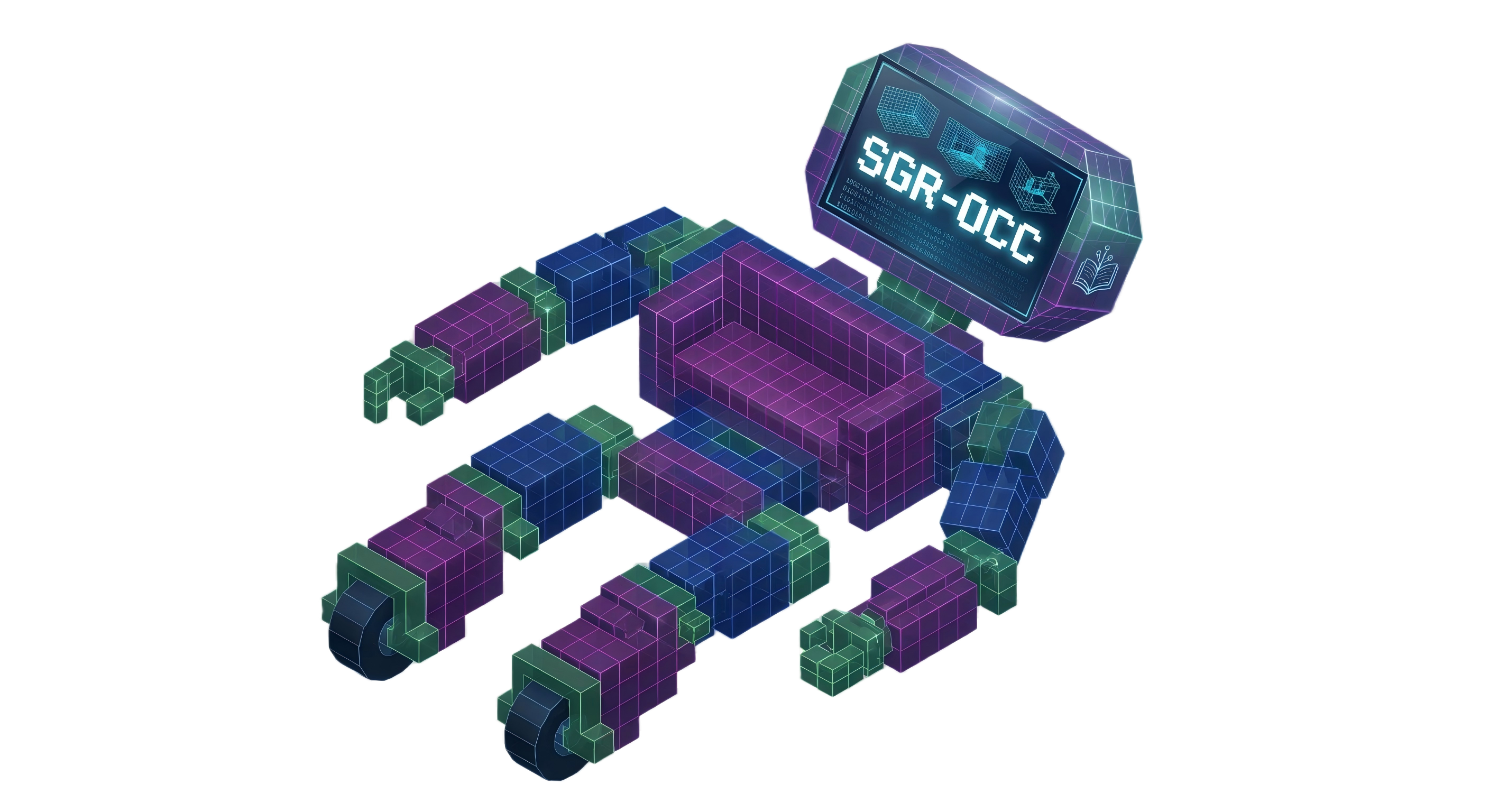}}}

\definecolor{bestcolor}{HTML}{E6F0FA}   
\definecolor{secondcolor}{HTML}{F5F9FD} 

\newcommand{\best}[1]{\cellcolor{bestcolor}\textbf{#1}}
\newcommand{\second}[1]{\cellcolor{secondcolor}\uline{#1}}

\definecolor{cCeiling}{HTML}{D9534F}  
\definecolor{cFloor}{HTML}{5CB85C}    
\definecolor{cWall}{HTML}{ADD8E6}     
\definecolor{cWindow}{HTML}{428BCA}   
\definecolor{cChair}{HTML}{F0AD4E}    
\definecolor{cBed}{HTML}{FFDAB9}      
\definecolor{cSofa}{HTML}{800080}     
\definecolor{cTable}{HTML}{00008B}    
\definecolor{cTvs}{HTML}{ADFF2F}      
\definecolor{cFurniture}{HTML}{FF8C00}
\definecolor{cObjects}{HTML}{D8BFD8}  

\usepackage[skip=10pt]{caption} 
\newtcolorbox{AIbox}[2][]{aibox,title=#2,#1}
\definecolor{lightblue}{rgb}{0.22,0.45,0.70}%
\definecolor{Gray}{gray}{0.95}
\definecolor{Cornsilk}{rgb}{1.0, 0.97, 0.86}

\usepackage{amsmath}

\usepackage[all]{hypcap}

\title{\paperlogo{} SGR-OCC: Evolving Monocular Priors for Embodied 3D Occupancy Prediction via Soft-Gating Lifting and Semantic-Adaptive Geometric Refinement}

\runningtitle{\paperlogo{} SGR-OCC: Evolving Monocular Priors for Embodied 3D Occupancy Prediction via Soft-Gating Lifting and Semantic-Adaptive Geometric Refinement}

\author{
  Yiran Guo$^1$,
  Simone Mentasti$^1$, 
  Xiaofeng Jin$^1$, 
  Matteo Frosi$^1$, and
  Matteo Matteucci$^1$
}

\affil[1]{Department of Electronics Information and Bioengineering, Politecnico di Milano}

\correspondingauthor{Matteo Matteucci \href{https://www.deib.polimi.it/eng/people/details/267262}{https://www.deib.polimi.it/eng/people/details/267262}; Email \href{matteo.matteucci@polimi.it}{matteo.matteucci@polimi.it}}

\begin{document}

\begin{abstract}
3D semantic occupancy prediction is a cornerstone for embodied AI, enabling agents to perceive dense scene geometry and semantics incrementally from monocular video streams. However, current online frameworks face two critical bottlenecks: the inherent depth ambiguity of monocular estimation that causes "feature bleeding" at object boundaries , and the "cold start" instability where uninitialized temporal fusion layers distort high-quality spatial priors during early training stages. In this paper, we propose SGR-OCC (Soft-Gating and Ray-refinement Occupancy), a unified framework driven by the philosophy of "Inheritance and Evolution". To perfectly inherit monocular spatial expertise, we introduce a Soft-Gating Feature Lifter that explicitly models depth uncertainty via a Gaussian gate to probabilistically suppress background noise. Furthermore, a Dynamic Ray-Constrained Anchor Refinement module simplifies complex 3D displacement searches into efficient 1D depth corrections along camera rays, ensuring sub-voxel adherence to physical surfaces. To ensure stable evolution toward temporal consistency, we employ a Two-Phase Progressive Training Strategy equipped with identity-initialized fusion, effectively resolving the cold start problem and shielding spatial priors from noisy early gradients. Extensive experiments on the EmbodiedOcc-ScanNet and Occ-ScanNet benchmarks demonstrate that SGR-OCC achieves state-of-the-art performance. In local prediction tasks, SGR-OCC achieves a completion IoU of 58.55$\%$ and a semantic mIoU of 49.89$\%$, surpassing the previous best method, EmbodiedOcc++, by 3.65$\%$ and 3.69$\%$ respectively. In challenging embodied prediction tasks, our model reaches 55.72$\%$ SC-IoU and 46.22$\%$ mIoU. Qualitative results further confirm our model's superior capability in preserving structural integrity and boundary sharpness in complex indoor environments.

\vspace{2mm}


\vspace{5mm}

\coloremojicode{1F4C5} \textbf{Date}: March 14, 2026


\github{} \textbf{Code Repository}: \href{https://github.com/Peking5t5/SGR-OCC}{https://github.com/Peking5t5/SGR-OCC}




\coloremojicode{1F4E7} \textbf{Contact}: \href{yiran.guo@polimi.it}{yiran.guo@polimi.it}, \href{matteo.matteucci@polimi.it}{matteo.matteucci@polimi.it}

\end{abstract}

\maketitle
\vspace{3mm}
\vspace{-4mm}
\begin{figure}[tb]
  \centering
  \includegraphics[width=\textwidth]{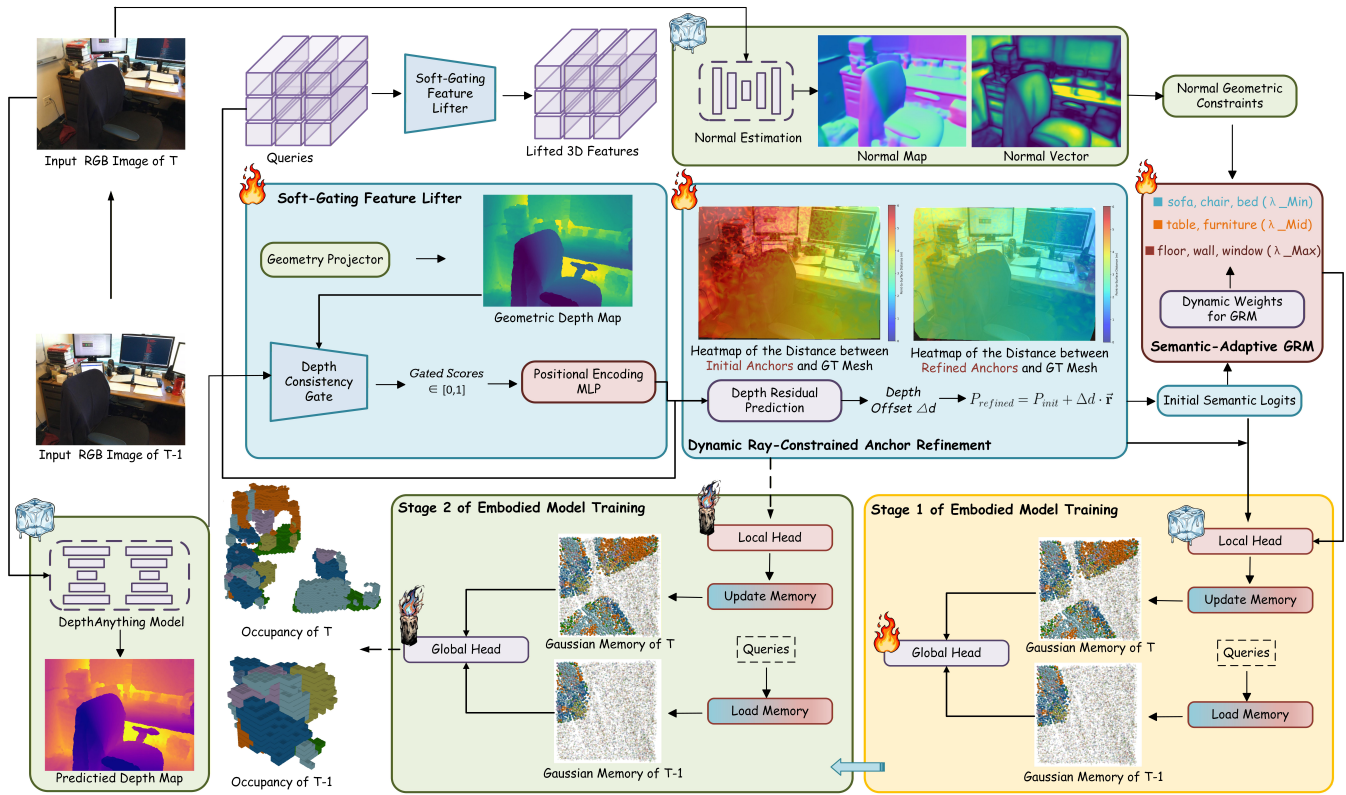}
  \caption{\textbf{The overall architecture of our proposed framework.} The pipeline consists of a Soft-Gating Feature Lifter for robust 2D-to-3D projection and a Dynamic Ray-Constrained Anchor Refinement module for sub-voxel geometric correction. The Semantic-Adaptive GRM (right) enforces category-specific geometric constraints. The bottom-right panel illustrates our Two-Phase Progressive Training Strategy: Stage 1 freezes the monocular backbone to initialize temporal alignment (training only the Global Head), while Stage 2 unfreezes the entire network for global co-adaptation using both Local and Global supervision.
  }
  \label{fig1}
\end{figure}

\section{Introduction}\label{sec:intro}

3D semantic occupancy prediction \cite{cao2022monoscenemonocular3dsemantic, huang2023tri, wei2023surroundocc, li2023voxformer, yu2023flashocc} has emerged as a fundamental representation for embodied AI and autonomous robotics, providing agents with a dense, unified understanding of both scene geometry and semantics. Unlike static, offline 3D reconstruction tasks that process entire video sequences simultaneously, embodied occupancy prediction requires an agent to incrementally build and update a global 3D map from a continuous stream of monocular observations \cite{wu2025embodiedocc, yu2024monocular, humblot2022navigation}. Achieving high-fidelity, real-time spatial awareness is critical for downstream interactive tasks such as obstacle avoidance, motion planning, and active exploration in complex indoor environments \cite{savva2019habitat, dai2017scannet, yu2024monocular, li2023neuralangelo, li2025sliceocc}.

Despite the rapid paradigm shift from offline volumetric methods to online streaming architectures \cite{li2023mseg3d}, current monocular-based frameworks still face two critical bottlenecks. First, the inherent ambiguity of monocular depth estimation severely limits spatial precision. Standard 2D-to-3D feature lifting operations \cite{philion2020lift, cao2022monoscenemonocular3dsemantic, li2024bevformer, huang2023tri} rely on deterministic depth values, leading to severe "feature bleeding" and feature misalignment at depth discontinuities (e.g., object boundaries). Second, transitioning from single-frame perception to multi-frame temporal fusion introduces the severe "cold start" instability \cite{wang2025embodiedocc++, ju2025video2bev}. When integrating temporal history, randomly initialized fusion layers produce noisy gradients that actively distort the high-quality spatial priors extracted by pre-trained backbones, causing a catastrophic "dilution" of geometric details during the early stages of training. 

To address the demand for efficient temporal accumulation, recent pioneering works, notably \textit{EmbodiedOcc} \cite{wu2025embodiedocc} and \textit{S2GO} \cite{park2025s2gostreamingsparsegaussian}, introduced the streaming Gaussian Memory mechanism \cite{kerbl20233d, huang2024gaussian}. By utilizing sparse Gaussian primitives to maintain historical observations, it provides a highly efficient foundation for online updates \cite{kong2025multi, keetha2024splatam}. However, while this mechanism excels at memory management, its performance ceiling is still bounded by the initial noise of monocular lifting and the gradient conflicts during temporal alignment \cite{wang2025embodiedocc++}. It necessitates a framework that can not only leverage this efficient memory but also robustly refine its geometric fidelity and temporal migration.

Motivated by this, we propose \textbf{SGR-OCC} (Soft-Gating and Ray-refinement Occupancy), a unified framework driven by the philosophy of \textit{"Inheritance and Evolution."} It comprises a \textbf{Spatial Expert} to perfectly inherit monocular spatial priors and mitigate depth ambiguity, and a \textbf{Temporal Manager} to smoothly evolve these priors into embodied temporal consistency without suffering from geometric degradation or fusion instability. 

\noindent In summary, our main contributions are four-fold:

\begin{itemize}
    \setlength{\topsep}{0pt}
    \setlength{\partopsep}{0pt}
    \setlength{\itemsep}{2pt}
    \setlength{\parsep}{0pt}
    \setlength{\parskip}{0pt}
    
    \item We propose \textbf{SGR-OCC}, an evolutionary framework that seamlessly integrates robust monocular spatial perception with stable temporal memory accumulation for embodied 3D occupancy prediction.
    \item We design a \textbf{Soft-Gating Feature Lifter} and a \textbf{Dynamic Ray-Constrained Refinement} module to physically resolve monocular depth ambiguities, ensuring boundary sharpness and sub-voxel accuracy.
    \item We introduce a \textbf{Two-Phase Progressive Training Strategy} and a \textbf{Semantic-Adaptive GRM} to effectively eliminate the "cold start" instability and preserve architectural planarity across dynamic updates.
    \item Extensive experiments show \textbf{SGR-OCC} achieves state-of-the-art results on the Occ-ScanNet and EmbodiedOcc-ScanNet \cite{wu2025embodiedocc} benchmarks, reaching 58.55\% and 55.72\% SC-IoU respectively, substantially outperforming existing baselines.
\end{itemize}



\vspace{-1mm}

\section{Related Work}
\label{sec:related}

\subsection{Vision-based 3D Occupancy Prediction}
Vision-based 3D semantic occupancy prediction aims to reconstruct a dense, voxelized representation of the 3D environment from 2D images. Early approaches heavily relied on depth estimation to lift 2D features into 3D space. The pioneering LSS \cite{philion2020lift} and its variants \cite{li2024bevformer, wei2025emd, wei2024nto3d} proposed frustum-based view transformation, which became the standard for bird's-eye-view (BEV) perception. Building upon this, MonoScene \cite{cao2022monoscenemonocular3dsemantic, wei2025omniindoor3d, zhao2024hybridocc} introduced the first framework for monocular 3D semantic scene completion using dense 3D convolutions. To mitigate the computational bottleneck of 3D grids, TPVFormer \cite{huang2023tri} proposed a tri-perspective view representation, while SurroundOcc \cite{wei2023surroundocc} utilized multi-scale 3D volume rendering. Recently, VoxFormer \cite{li2023voxformer}, FlashOcc \cite{yu2023flashocc} and OccRWKV \cite{wang2025occrwkv} introduced sparse voxel transformers and channel-to-height plugins to further accelerate the prediction pipeline. Despite these advancements, most existing methods are designed for offline, static scenes and rely on deterministic 2D-to-3D hard-projection \cite{cao2022monoscenemonocular3dsemantic, huang2023tri, zuo2023pointocc, zhang2023occformer}. They struggle with the inherent depth ambiguity of monocular vision, often resulting in severe "feature bleeding" at object boundaries. In contrast, our \textbf{SGR-OCC} introduces a \textit{Soft-Gating Feature Lifter} that probabilistically models depth uncertainty, coupled with a \textit{Dynamic Ray-Constrained Refinement} that physically restricts anchor optimization to 1D camera rays ($\mathbb{R}^1$), significantly enhancing sub-voxel accuracy.

\subsection{Embodied and Online Scene Understanding}
Unlike offline 3D reconstruction that processes entire sequences simultaneously, embodied AI requires agents to incrementally perceive and navigate through streaming observations. Recent autonomous driving frameworks like StreamPETR \cite{wang2023exploring}, BEVCon \cite{leng2025bevcon} and GDFusion \cite{chen2025rethinking} have demonstrated the potential of recurrent architectures for temporal fusion. Pushing this to indoor robotics, EmbodiedOcc \cite{wu2025embodiedocc} defined the embodied 3D occupancy prediction task, highlighting the need for real-time, memory-efficient global scene updates. Its successor, EmbodiedOcc++ \cite{wang2025embodiedocc++}, further integrated plane regularization and uncertainty sampling to refine architectural structures. A major bottleneck in these online frameworks is the "cold start" instability during temporal migration \cite{wang2025embodiedocc++}. Randomly initialized recurrent fusion layers tend to actively distort the pre-trained spatial monocular features in early training epochs. Moreover, uniform geometric regularization often over-smooths fine-grained objects \cite{tang2024sparseocc, zhang2023occformer, wang2024embodiedscan, sun2021neuralrecon}. Our framework resolves these issues via a \textit{Two-Phase Progressive Training Strategy} featuring identity-initialized fusion, effectively shielding spatial priors. Additionally, we propose a \textit{Semantic-Adaptive GRM} that dynamically restricts planar constraints to structural categories (e.g., walls) while preserving complex object geometries.

\subsection{Gaussian-based 3D Representations}
The advent of 3D Gaussian Splatting (3DGS) \cite{kerbl20233d, wang2024embodiedscan, Wei2024GraphAvatarCH} has revolutionized 3D scene representation by replacing dense grids with explicit, continuous Gaussian primitives, enabling real-time rendering and optimization \cite{}. Inspired by this, GaussianFormer \cite{huang2024gaussian} and GaussianOcc3D \cite{doruk2026gaussianocc3d} modeled the 3D scene as discrete Gaussians for occupancy prediction. To handle infinite streaming data, \textit{S2GO} \cite{park2025s2gostreamingsparsegaussian}, \textit{EmbodiedOcc} \cite{wu2025embodiedocc} and GaussianWorld \cite{zuo2025gaussianworld} introduced the Gaussian Memory mechanism, which efficiently accumulates historical primitives in an online fashion without unbounded memory growth. While Gaussian Memory provides an elegant solution for temporal storage, directly trusting unconstrained monocular Gaussian initializations leads to spatial drift over long trajectories \cite{turkulainen2025dn, yu2024gsdf, guedon2024sugar, chen2024pgsr, zhang2024rade}. Our framework \textit{inherits} this efficient Gaussian Memory but \textit{evolves} it by introducing a hybrid confidence-driven verification mechanism. By jointly evaluating geometric reprojection consistency and calibrated semantic probability, our temporal manager meticulously manages the lifecycle of each Gaussian primitive, preventing ghosting artifacts and ensuring long-term spatial-temporal consistency.
\section{Methodology}
\subsection{Framework Overview}\label{sec3.1}
We present a unified framework for high-fidelity 3D semantic occupancy prediction from monocular image sequences. Given a sequence of images $\mathcal{I} = \{I_1, \dots, I_T\}$ and their corresponding camera poses $\mathcal{P} = \{P_1, \dots, P_T\}$, our goal is to reconstruct a semantically consistent global 3D scene representation $O_{global}$. Driven by the core philosophy of "Inheritance and Evolution," the proposed architecture (illustrated in Fig. \ref{fig1}) consists of two complementary stages and a tailored training strategy:

\textbf{Stage 1: High-Fidelity Spatial Perception (The Spatial Expert).} This stage functions as the geometric foundation to generate precise 3D primitives from 2D observations. It mitigates inherent monocular depth ambiguity using a Soft-Gating Feature Lifter (Sec. \ref{sec3.2}) and enforces strict geometric consistency via a Dynamic Ray-Constrained Anchor Refinement module (Sec. \ref{sec3.3}). This ensures sub-voxel accuracy and effectively prevents the "garbage-in" problem before temporal fusion.

\textbf{Stage 2: Robust Temporal Evolution (The Temporal Manager).} Instead of simple concatenation, this stage fuses sequential information into a coherent global map by inheriting pre-trained spatial weights. It aggregates "Gaussian Memory" across frames and applies a Semantic-Adaptive Geometric Regularization (GRM) (Sec. \ref{sec3.4}) to the global output, enforcing category-specific geometric constraints (e.g., planarity for walls) to keep the accumulated scene structurally rigorous.

\textbf{Two-Phase Progressive Training.} To bridge the domain gap between monocular priors and embodied consistency, we adopt a Two-Phase Progressive Training Strategy (Sec. \ref{sec3.5}). This progressive schedule strictly isolates temporal alignment—where spatial components are frozen to solve the "cold start" instability—from global co-adaptation, which unfreezes the entire network using differential learning rates for joint optimization.

\subsection{High-Fidelity Spatial Perception via Soft-Gating (The Spatial Expert)}\label{sec3.2}
To serve as a robust foundation for the embodied system, the spatial perception module must generate high-quality 3D features from single-view images, handling inherent depth ambiguities at object boundaries. We propose a Soft-Gating Feature Lifter that inherently models depth uncertainty to guide the 2D-to-3D projection.

\textbf{Depth-Aware Deformable Aggregation.}\label{sec3.2.1} Standard hard-projection methods lift 2D features based on a deterministic depth value, which often leads to feature misalignment at depth discontinuities (e.g., object edges) \cite{philion2020lift, li2024bevformer}. Instead, we formulate the lifting process as a probabilistic aggregation within a deformable window.
For a 3D query point $\mathbf{q} \in \mathbb{R}^3$, we project it onto the 2D image plane to obtain the reference point $\mathbf{p}_{ref}$ and its projected depth $d_{proj}$ \cite{yang2024depth, lin2025depth, reading2021categorical}. We then learn $K$ sampling offsets $\{\Delta \mathbf{p}_k\}_{k=1}^K$ around $\mathbf{p}_{ref}$. The lifted 3D feature $\mathbf{F}_{3D}$ is computed as Equation \ref{eq1}:

\begin{equation}
\label{eq1}
    \mathbf{F}_{3D}(\mathbf{q}) = \sum_{k=1}^{K} A_{k} \cdot \mathcal{G}(d_{proj}, d_{pred}^{(k)}) \cdot \mathbf{F}_{2D}(\mathbf{p}_{ref} + \Delta \mathbf{p}_k)
\end{equation}

where $A_k$ is the learned attention weight, and $\mathbf{F}_{2D}$ is the image feature map. Crucially, $\mathcal{G}(\cdot)$ is our proposed Gaussian Depth Gate, which weighs the contribution of each sampling point based on depth consistency.

\begin{figure}[tb]
  \centering
  \begin{subfigure}{0.54\linewidth}
    \centering
    \includegraphics[width=\linewidth]{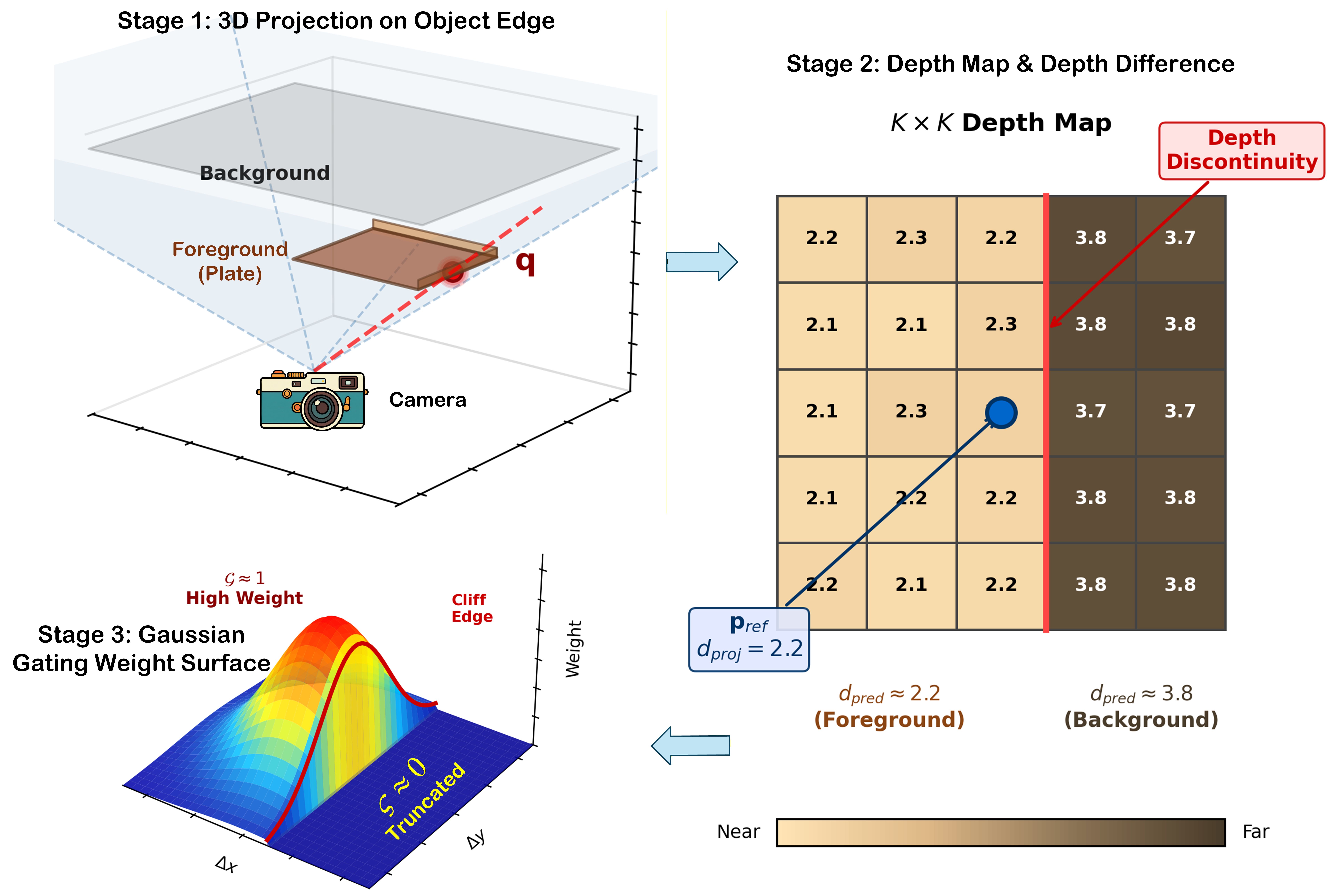} 
    \caption{\textbf{Soft-Gating Feature Lifter}}
    \label{fig:2-a}
  \end{subfigure}
  \hfill
  \begin{subfigure}{0.42\linewidth} 
    \centering
    \includegraphics[width=\linewidth]{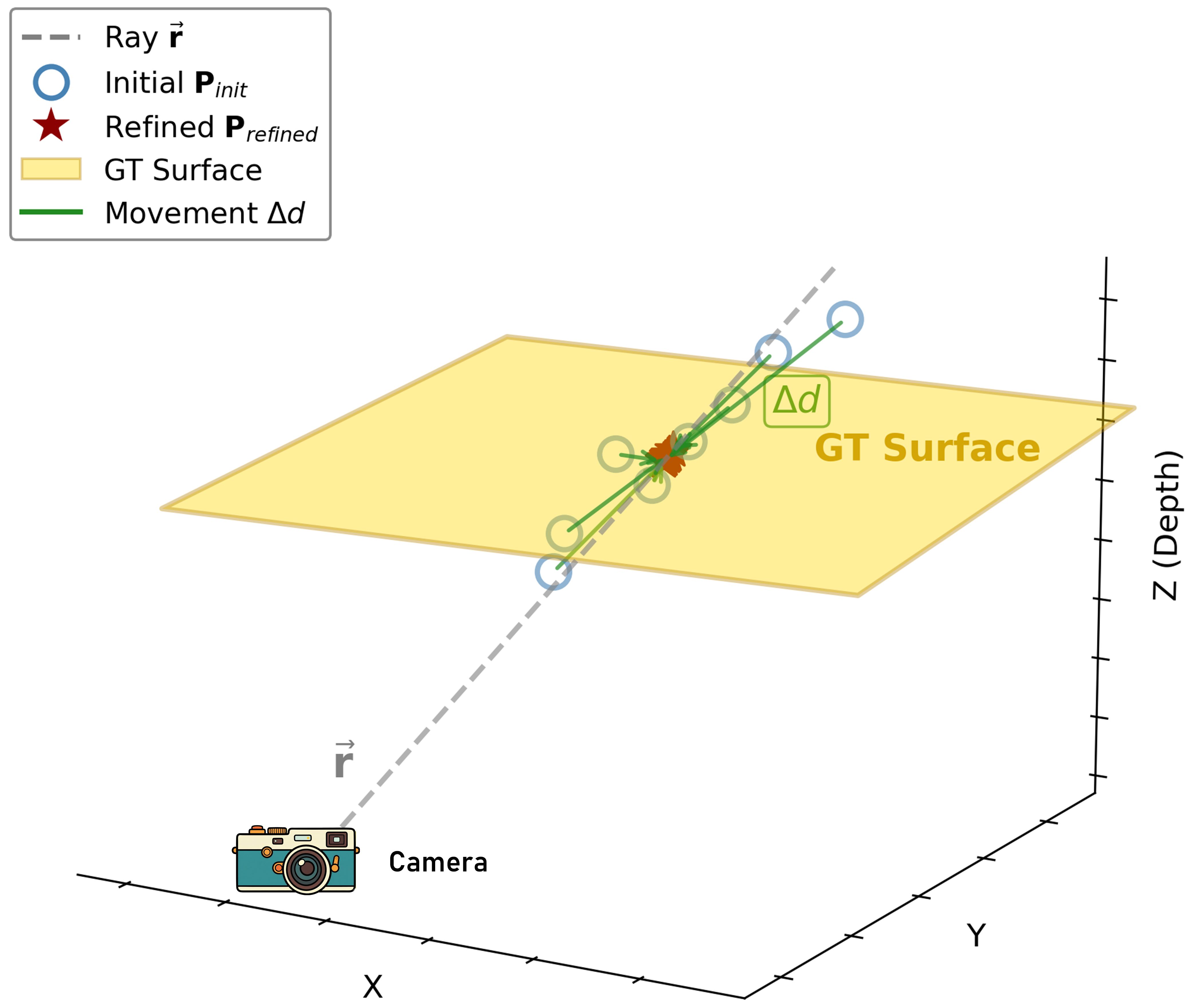}
    \caption{\textbf{Ray-Constrained Refinement}}
    \label{fig:2-b}
  \end{subfigure}
  
  \caption{\textbf{Overview of the core geometric modules.} \textbf{(a)} The Soft-Gating mechanism evaluates depth consistency ($d_{proj}$ vs. $d_{pred}$) within a sampling window. It applies Gaussian weights to foreground regions while truncating conflicting background noise, ensuring robust feature aggregation. \textbf{(b)} The Refinement module restricts anchor optimization ($\mathbf{P}_{init} \to \mathbf{P}_{refined}$) to a 1D depth residual $\Delta d$ strictly along the camera ray $\vec{\mathbf{r}}$, adhering points to the physical surface and converting a complex 3D search into an efficient 1D correction.}
  \label{fig2}
\end{figure}

\textbf{Physics-Based Gaussian Gating.}\label{sec3.2.2} 
Recognizing that monocular depth estimates are inherently stochastic, we reformulate the feature lifting process through the lens of geometric adherence \cite{cao2022monoscenemonocular3dsemantic}. Unlike typical MLP-based attention which lacks physical grounding, we interpret the depth consistency as a spatial probability density function (PDF) \cite{qiu2025gated}. The gating score $G_k$ measures the likelihood of a sampling point residing on the physical surface, defined by a learnable Gaussian kernel:
\begin{equation}
\label{eq2}
    \mathcal{G}(d_{proj}, d_{pred}^{(k)}) = \alpha \cdot \exp\left( - \frac{\| d_{proj} - d_{pred}^{(k)} \|^2}{2\sigma^2} \right)
\end{equation}
where $d_{pred}^{(k)}$ is the predicted depth, with $\alpha$ and $\sigma$ serving as learnable scale and tolerance parameters, respectively. Theoretically, this formulation acts as a soft-constraint in the kernel space; it assigns maximum density to points with high geometric agreement while exponentially suppressing "off-surface" noise. As Fig. \ref{fig:2-a} illustrates, this mechanism provides a differentiable manifold to resolve the "feature bleeding" problem at object boundaries. By probabilistically filtering depth-inconsistent background features, we ensure that only geometrically valid representations contribute to the final voxel occupancy, maintaining a continuous gradient flow that is superior to discrete, hard-thresholding alternatives.

\textbf{Dynamic Ray-Constrained Anchor Refinement.}\label{sec3.2.3} To further strictly enforce geometric validity, we constrain the optimization of 3D anchors. In monocular reconstruction, geometric uncertainty is predominantly distributed along the camera ray rather than the image plane \cite{mildenhall2021nerf, mescheder2019occupancy, li2023voxformer, yu2023flashocc}. Therefore, for an initial anchor position $\mathbf{P}_{init}$, we predict a scalar depth residual $\Delta d$ instead of a free 3D vector as Figure \ref{fig:2-b} showing. The refined position $\mathbf{P}_{refined}$ is constrained as Equation \ref{eq3}:

\begin{equation}
\label{eq3}
    \mathbf{P}_{refined} = \mathbf{P}_{init} + \Delta d \cdot \frac{\mathbf{P}_{init} - \mathbf{O}_{cam}}{\| \mathbf{P}_{init} - \mathbf{O}_{cam} \|}
\end{equation}

where $\mathbf{O}_{cam}$ is the camera center. This Ray-Constraint effectively reduces the search space from $\mathbb{R}^3$ to $\mathbb{R}^1$, ensuring sub-voxel accuracy and preventing the optimization from drifting into physically implausible configurations.

\subsection{Robust Temporal Evolution via Inheritance (The Temporal Manager)}\label{sec3.3}

While the spatial expert provides high-fidelity snapshots, the embodied system must accumulate these observations into a coherent global representation without introducing noise or catastrophic forgetting. We propose an Inheritance-Based Temporal Architecture that seamlessly evolves the monocular prior into a temporal estimator.

\textbf{Visibility-Aware Gaussian Pool Management.}\label{sec3.3.1} To manage the lifecycle of geometric primitives over long sequences, we maintain a global Gaussian Pool $\mathcal{M}_t$. Each primitive $g_i \in \mathcal{M}_t$ is assigned a semantic tag $\tau_i \in \{0, 1\}$ distinguishing its source: $\tau_i=1.0$ for high-confidence predicted anchors, and $\tau_i=0.0$ for newly initialized random ones \cite{kerbl20233d, huang2024gaussian, park2025s2gostreamingsparsegaussian}. At each time step $t$, we perform a visibility check using the current camera pose $\mathbf{T}_t$. The pool is updated by retaining historical anchors that are visible and geometrically consistent, while discarding occluded or invalid random anchors \cite{wu2025embodiedocc, wang2025embodiedocc++}. This Tag-Based Management ensures that the encoder processes a clean mixture of "verified history" and "new observations," rather than being overwhelmed by accumulated noise.

\textbf{Identity-Initialized Temporal Fusion.}\label{sec3.3.2} A major challenge in training embodied models is the "cold start" problem, where an uninitialized temporal fusion layer distorts the high-quality features from the pre-trained spatial encoder. To resolve this, we introduce an Identity-Initialized Fusion strategy.

Let $\mathbf{F}_{curr}$ be the feature form the current frame (Stage 1 output) and $\mathbf{F}_{hist}$ be the retrieved historical feature. The fused feature $\mathbf{F}_{fused}$ is computed via a residual recurrent block:

\begin{equation}
\label{eq4}
    \mathbf{F}_{fused} = \text{Fusion}(\mathbf{F}_{hist}, \mathbf{F}_{curr}) = \mathbf{W}_{h} \mathbf{F}_{hist} + \mathbf{W}_{c} \mathbf{F}_{curr} + \mathbf{b}
\end{equation}

Crucially, we initialize the weights $\mathbf{W}_{h}$ and $\mathbf{W}_{c}$ to explicitly enforce an identity mapping at the start of training, as Equation \ref{eq5}:
\begin{equation}
\label{eq5}
    \mathbf{W}_{h}^{init} = \mathbf{0}, \quad \mathbf{W}_{c}^{init} = \mathbf{I}, \quad \mathbf{b}^{init} = \mathbf{0}
\end{equation}

Under this initialization, at epoch 0, $\mathbf{F}_{fused} \equiv \mathbf{F}_{curr}$. This guarantees that the embodied model behaves exactly like the fully trained spatial expert initially. As training progresses, the model gradually learns non-zero weights for $\mathbf{W}_{h}$ to incorporate temporal context only when it provides information gain \cite{liu2019planercnn, huang2023tri}. This strategy effectively eliminates the performance drop typically seen in the early stages of embodied transfer learning.

\subsection{Hybrid Confidence-Driven Geometry Refinement Mechanism}\label{sec3.4}
To handle dynamic objects and occlusions without ghosting artifacts or temporal flickering, we propose a Hybrid Confidence-Driven Geometry Refinement Mechanism based on geometric consistency and calibrated semantics.

\textbf{Geometric Consistency via Depth Reprojection.}\label{sec3.4.1} The first verification checks physical validity: Does the historical anchor physically exist in the current view?
We leverage the depth maps $D_{pred}$ predicted by our Soft-Gating Lifter (Sec. \ref{sec3.2}) as the geometric ground truth for the current frame. For each reused anchor $P_{t-1}$, we reproject it to the current image plane $(u, v)$ to obtain its projected depth $d_{proj}$. The geometric confidence is modeled using a Gaussian kernel on the depth residual, as Equation \ref{eq6} saying:

\begin{equation}
\label{eq6}
    C_{geo} = \exp\left( - \frac{\| d_{proj} - D_{pred}(u, v) \|^2}{2\sigma_{geo}^2} \right)
\end{equation}

where $\sigma_{geo}$ controls the tolerance for depth mismatch. A high $C_{geo}$ indicates that the anchor is consistent with the current geometry, while a low score suggests occlusion or dynamic movement, triggering a position update.

\textbf{Calibrated Semantic Probability.}\label{sec3.4.2} The second verification checks cognitive stability: Is the semantic identity of the anchor consistent? Raw softmax probabilities from neural networks are often uncalibrated and over-smoothed (e.g., maximum probability $\approx 0.5$), which suppresses the confidence of correct predictions. To address this, we apply Temperature Scaling and Thresholding to calibrate the semantic confidence:

\begin{equation}
\label{eq7}
    P_{calib} = \text{Softmax}(\frac{\mathbf{l}_{sem}}{T}), \quad C_{sem} = \text{Clamp}\left(\frac{\max(P_{calib}) - \tau_{min}}{\tau_{max} - \tau_{min}}, 0, 1\right)
\end{equation}

where $\mathbf{l}_{sem}$ are the semantic logits, $T$ is the temperature coefficient (default $T=0.5$) to sharpen the distribution, and $[\tau_{min}, \tau_{max}]$ defines the confidence range. This calibration ensures that ambiguous predictions (low confidence) allow for refinement, while distinct predictions (high confidence) enforce stability.

\textbf{Unified Update Strategy.}\label{sec3.4.3} The final Hybrid Confidence $\mathcal{C}_{final}$ is computed as the product of the tag-based mask and the dual verification scores, as Equation \ref{eq8} represents:

\begin{equation}
\label{eq8}
    \mathcal{C}_{final} = \underbrace{\mathbb{I}_{old}}_{\text{Identity Protection}} \cdot \underbrace{(C_{geo} \cdot C_{sem})}_{\text{Dual Verification}}
\end{equation}

where $\mathbb{I}_{old}$ is a binary indicator (1 for reused anchors, 0 for new anchors). We apply a differential update ratio $\lambda = 1 - \mathcal{C}_{final}$: new anchors ($\lambda=1.0$) are fully updated via ray-constrained refinement, verified old anchors ($\lambda \approx 0.0$) are frozen for stability, and conflicting anchors ($\lambda > 0.5$) are significantly updated or re-initialized.

\subsection{Two-Phase Training Strategy}\label{sec3.5}
To achieve robust scene-level occupancy prediction in embodied environments, we propose a Two-Phase Training Strategy combined with a Dual-Head Architecture.
\textbf{Dual-Head Architecture: Local and Global Supervision.}\label{sec3.5.1} As illustrated in Fig. \ref{fig1}, our model employs two specialized prediction heads to handle different spatial-temporal scales: Local Head: This head operates on a per-frame basis with a resolution of $H=60, W=60, D=36$. Global Head predicts the complete scene occupancy with a larger volume ($H=200, W=220, D=90$). To ensure semantic alignment between the two heads, we implement a weight-sharing mechanism for fundamental categories, allowing the global head to inherit the refined discriminative power of the local head.
\textbf{Phase 2: Global Co-adaptation via Differential Fine-tuning.}\label{sec3.5.2} In the subsequent global co-adaptation phase (the last 10 epochs), we unfreeze the network and perform end-to-end optimization \cite{pan2025idinit}. However, applying a uniform learning rate across all modules would inadvertently destroy the well-established spatial priors. To prevent this, we propose a Hierarchical Differential Learning Rate strategy to prevent destroying spatial priors: \textbf{(1) Strict Preservation:} The 2D vision backbone remains frozen; \textbf{(2) Conservative Adaptation:} The Spatial Expert is fine-tuned with a significantly decayed learning rate to preserve soft-gating behaviors; \textbf{(3) Active Alignment:} The Temporal Manager and heads receive the full base learning rate to rapidly establish spatio-temporal consistency.

\vspace{-2mm}
\section{Experiments}\label{sec4}
\subsection{Experimental Setup \& Implementation Details}\label{sec4.1}

\textbf{Datasets and Tasks.} We evaluate our framework on the large-scale Occ-ScanNet \cite{yu2024monocular} and its sequential extension EmbodiedOcc-ScanNet \cite{wu2025embodiedocc} benchmark (537 train / 137 val scenes, 30 posed frames per sequence). The semantic space comprises 12 distinct classes plus an empty class. We conduct evaluations on two complementary tasks: \textbf{Local Occupancy Prediction}, which predicts a $60 \times 60 \times 36$ voxel grid (0.08m resolution) within the camera frustum, and \textbf{Embodied Occupancy Prediction}, which dynamically reconstructs the global scene dimensions via online temporal integration. 

\textbf{Evaluation Metrics.} Following established protocols \cite{wu2025embodiedocc}, we quantitatively assess performance using Scene Completion Intersection over Union (SC-IoU) to evaluate overall 3D geometric accuracy, and semantic mean Intersection over Union (mIoU) to measure fine-grained categorical reconstruction. 

\textbf{Implementation Details.} Our framework incorporates a pre-trained Depth-Anything-V2 \cite{yang2024depth} model as the frozen vision backbone. Key hyperparameter configurations include the Soft-Gating depth tolerance $\sigma=0.5$, temperature scaling $T=0.5$, and temporal confidence bounds $[\tau_{min}, \tau_{max}]=[0.2, 0.8]$. The Semantic-Adaptive GRM dynamically applies strong planar constraints (weight $\approx 1.0$) to structural categories while relaxing them for object-centric ones. Networks are optimized using AdamW (weight decay 0.01) with a cosine annealing schedule on 8 NVIDIA A6000 GPUs. 

\textbf{Optimization Strategy.} For local occupancy, we train the Spatial Expert for 10 epochs (max LR (Learning-Rate): $2\times10^{-4}$). For embodied occupancy, we strictly execute our Two-Phase Progressive Training over 15 epochs. In Phase 1 (Epochs 1-5), the spatial components are frozen, and the Temporal Manager is trained (LR $1\times10^{-4}$) using Identity-Initialized Fusion ($W_h=0, W_c=I$) to guarantee a stable cold start. In Phase 2 (Epochs 6-15), we unfreeze the network for global co-adaptation using differential learning rates: the 2D backbone remains completely frozen ($0\times$), the lifter and spatial encoders use a conservative multiplier ($0.1\times$, i.e., $1\times10^{-5}$), and the temporal fusion modules maintain the base LR ($1.0\times$, i.e., $1\times10^{-4}$). \textit{Extensive details regarding architectures and parameters are provided in the Supplementary Material.}

\subsection{Main Results}\label{sec4.3}
\textbf{Local Occupancy Prediction.} Tab.~\ref{tab1} summarizes the quantitative results on the Occ-ScanNet dataset. SGR-OCC achieves state-of-the-art performance across all primary metrics, reaching 58.55$\%$ SC-IoU and 49.89$\%$ mIoU, which outperforms the previous best method, EmbodiedOcc++ \cite{wang2025embodiedocc++}, by +3.65$\%$ and +3.69$\%$ respectively. Furthermore, our method shows a massive >16$\%$ SC-IoU improvement over early volumetric paradigms \cite{cao2022monoscenemonocular3dsemantic, huang2023tri}. Category-wise, our model excels in both large structural elements (Wall: 54.50$\%$, Window: 50.80$\%$) and fine-grained indoor objects (Furniture: 56.70$\%$, TVs: 44.30$\%$), demonstrating that our architecture preserves complex geometric details alongside scene layouts.

\begin{table}
\centering
\caption{Local Prediction Performance on the Occ-ScanNet dataset.}
\label{tab1}
\resizebox{\textwidth}{!}{%
\setlength{\tabcolsep}{4pt} 
\begin{tabular}{l|c|c|c|c|c|c|c|c|c|c|c|c|c|c|c}
\hline
Method & Input & SC-IoU &
\rotatebox{90}{\textcolor{cCeiling}{$\blacksquare$}~ceiling} & 
\rotatebox{90}{\textcolor{cFloor}{$\blacksquare$}~floor} & 
\rotatebox{90}{\textcolor{cWall}{$\blacksquare$}~wall} & 
\rotatebox{90}{\textcolor{cWindow}{$\blacksquare$}~window} & 
\rotatebox{90}{\textcolor{cChair}{$\blacksquare$}~chair} & 
\rotatebox{90}{\textcolor{cBed}{$\blacksquare$}~bed} & 
\rotatebox{90}{\textcolor{cSofa}{$\blacksquare$}~sofa} & 
\rotatebox{90}{\textcolor{cTable}{$\blacksquare$}~table} & 
\rotatebox{90}{\textcolor{cTvs}{$\blacksquare$}~tvs} & 
\rotatebox{90}{\textcolor{cFurniture}{$\blacksquare$}~furniture} & 
\rotatebox{90}{\textcolor{cObjects}{$\blacksquare$}~objects} & 
mIoU \\
\hline 

TPVFormer (\cite{huang2023tri}) & $x^{\text {rgb }}$ & 33.39 & 6.96 & 32.97 & 14.41 & 9.10 & 24.01 & 41.49 & 45.44 & 28.61 & 10.66 & 35.37 & 25.31 & 24.94 \\

GaussianFormer (\cite{huang2024gaussian}) & $x^{\text {rgb }}$ & 40.91 & 20.70 & 42.00 & 23.40 & 17.40 & 27.0 & 44.30 & 44.80 & 32.70 & 15.30 & 36.70 & 25.00 & 29.93 \\

MonoScene (\cite{cao2022monoscenemonocular3dsemantic}) & $x^{\text {rgb }}$ & 41.60 & 15.17 & 44.71 & 22.41 & 12.55 & 26.11 & 27.03 & 35.91 & 28.32 & 6.57 & 32.16 & 19.84 & 24.62 \\
 
ISO (\cite{yu2024monocular}) & $x^{\text {rgb }}$ & 42.16 & 19.88 & 41.88 & 22.37 & 16.98 & 29.09 & 42.43 & 42.00 & 29.60 & 10.62 & 36.36 & 24.61 & 28.71 \\
 
Surroundocc (\cite{wei2023surroundocc}) & $x^{\text {rgb }}$ & 42.52 & 18.90 & 49.30 & 24.80 & 18.00 & 26.80 & 42.00 & 44.10 & 32.90 & 18.60 & 36.80 & 26.90 & 30.83 \\

EmbodiedOcc (\cite{park2025s2gostreamingsparsegaussian}) & $x^{\mathrm{rgb}}$ & 53.55 & \second{39.60} & 50.40 & 41.40 & 31.70 & 40.90 & 55.00 & 61.40 & 44.00 & \second{36.10} & 53.90 & 42.20 & 45.15 \\

EmbodiedOcc++ (\cite{wang2025embodiedocc++}) & $x^{\mathrm{rgb}}$ & \second{54.90} & 36.40 & \second{53.10} & \second{41.80} & \second{34.40} & \second{42.90} & \second{57.30} & \best{64.10} & \second{45.20} & 34.80 & \second{54.20} & \second{44.10} & \second{46.20} \\

\textbf{SGR-OCC (Ours)} & $x^{\mathrm{rgb}}$ & \best{58.55} & \best{44.30} & \best{54.50} & \best{50.80} & \best{40.10} & \best{44.40} & \best{57.60} & 63.60 & \best{46.40} & \best{44.30} & \best{56.70} & \best{46.10} & \best{49.89} \\
\hline
\end{tabular}
}
\end{table}

\begin{figure}[t]
  \centering
  \includegraphics[width=\linewidth]{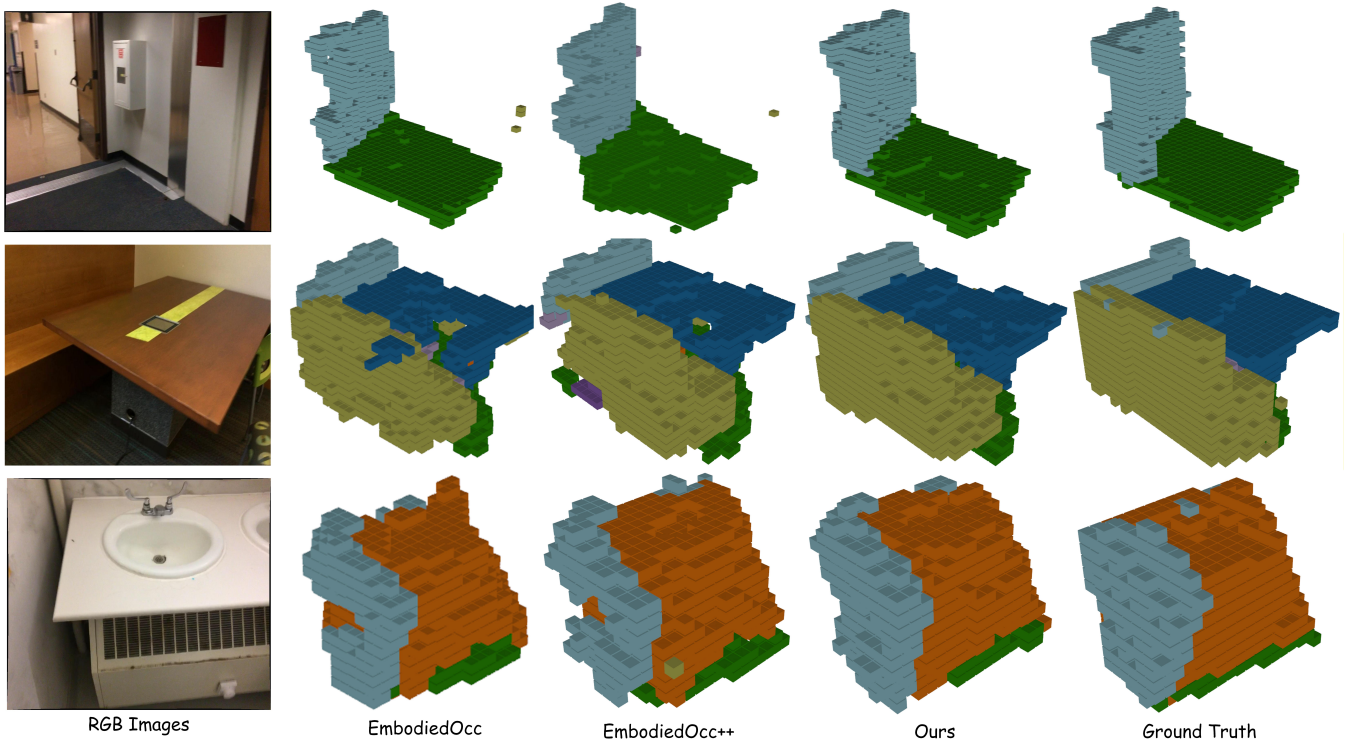}
  \caption{\textbf{Qualitative comparison of local occupancy prediction on Occ-ScanNet.} From left to right: Input RGB images, EmbodiedOcc \cite{wu2025embodiedocc}, EmbodiedOcc++ \cite{wang2025embodiedocc++}, our proposed method, and the Ground Truth. Our approach exhibits superior boundary sharpness and structural integrity, particularly in complex geometries like the sink (row 3) and table edges (row 2), thanks to the Soft-Gating mechanism and Ray-Constrained Refinement.}
  \label{fig3}
\end{figure}

\textbf{Qualitative Results (Local).} As visually confirmed in Fig.~\ref{fig3}, our method produces significantly cleaner geometric structures than current SOTA models. While EmbodiedOcc \cite{wu2025embodiedocc} and EmbodiedOcc++ \cite{wang2025embodiedocc++} suffer from severe "bleeding" artifacts at object boundaries (table edges in row 2) and physically implausible floating voxels (sink in row 3), SGR-OCC yields sharp, well-defined contours. This demonstrates the direct efficacy of our geometric modules in suppressing depth noise and strictly adhering anchors to physical surfaces.

\textbf{Embodied Occupancy Prediction.} To assess spatio-temporal integration, we evaluate global scene prediction after processing full 30-frame sequences (Tab.~\ref{tab2}). SGR-OCC consistently dominates all baselines, achieving 55.72$\%$ SC-IoU and 46.22$\%$ mIoU (outperforming EmbodiedOcc++ by +3.52$\%$ and +2.62$\%$). This superiority is evident across both structural categories (Ceiling: +11.7$\%$, Wall: +10.1$\%$) and object-centric categories (Table: +8.0$\%$, Objects: +9.2$\%$). These massive category-level gains prove that our framework effectively overcomes the "cold start" dilution and reduces the error accumulation.

\begin{table}
\centering
\caption{Embodied prediction performance on the EmbodiedOcc-ScanNet dataset.}
\label{tab2}
\resizebox{\textwidth}{!}{%
\setlength{\tabcolsep}{4pt} 
\begin{tabular}{l|c|c|c|c|c|c|c|c|c|c|c|c|c|c|c}
\hline
Method & Dataset & SC-IoU &
\rotatebox{90}{\textcolor{cCeiling}{$\blacksquare$}~ceiling} & 
\rotatebox{90}{\textcolor{cFloor}{$\blacksquare$}~floor} & 
\rotatebox{90}{\textcolor{cWall}{$\blacksquare$}~wall} & 
\rotatebox{90}{\textcolor{cWindow}{$\blacksquare$}~window} & 
\rotatebox{90}{\textcolor{cChair}{$\blacksquare$}~chair} & 
\rotatebox{90}{\textcolor{cBed}{$\blacksquare$}~bed} & 
\rotatebox{90}{\textcolor{cSofa}{$\blacksquare$}~sofa} & 
\rotatebox{90}{\textcolor{cTable}{$\blacksquare$}~table} & 
\rotatebox{90}{\textcolor{cTvs}{$\blacksquare$}~tvs} & 
\rotatebox{90}{\textcolor{cFurniture}{$\blacksquare$}~furniture} & 
\rotatebox{90}{\textcolor{cObjects}{$\blacksquare$}~objects} & 
mIoU \\
\hline 

TPVFormer (\cite{huang2023tri}) & EmbodiedOcc & 35.88 & 1.62 & 30.54 & 12.03 & 13.22 & 35.47 & 51.39 & 49.79 & 25.63 & 3.60 & 43.15 & 16.23 & 25.70 \\

SurroundOcc (\cite{wei2023surroundocc}) & EmbodiedOcc & 37.04 & 12.70 & 31.80 & 22.50 & 22.00 & 29.90 & 44.70 & 36.50 & 24.60 & 11.50 & 34.40 & 18.20 & 26.27 \\

GaussianFormer (\cite{huang2024gaussian}) & EmbodiedOcc & 38.02 & 17.00 & 33.60 & 21.50 & 21.70 & 29.40 & 47.80 & 37.10 & 24.30 & 15.50 & 36.20 & 16.80 & 27.36 \\
 
SplicingOcc & EmbodiedOcc & 49.01 & \second{31.60} & 38.80 & 35.50 & 36.30 & 47.10 & 54.50 & 57.20 & 34.40 & 32.50 & 51.20 & 29.10 & 40.74 \\

EmbodiedOcc (\cite{park2025s2gostreamingsparsegaussian}) & EmbodiedOcc & 51.52 & 22.70 & \second{44.60} & 37.40 & \second{38.00} & \best{50.10} & \second{56.70} & \best{59.70} & 35.40 & \second{38.40} & 52.00 & 32.90 & 42.53 \\

EmbodiedOcc++ (\cite{wang2025embodiedocc++}) & EmbodiedOcc & \second{52.20} & 27.90 & 43.90 & \second{38.70} & \best{40.60} & \second{49.00} & \best{57.90} & \second{59.20} & \second{36.80} & 37.80 & \best{53.50} & \second{34.10} & \second{43.60} \\

\textbf{SGR-OCC (Ours)} & EmbodiedOcc & \best{55.72} & \best{39.60} & \best{52.50} & \best{
48.80} & 34.30 & 42.90 & 51.10 & 58.50 & \best{44.80} & \best{40.30} & \second{52.40} & \best{43.30} & \best{46.22} \\
\hline
\end{tabular}
}
\end{table}

\begin{figure}[t]
  \centering
  \includegraphics[width=\linewidth]{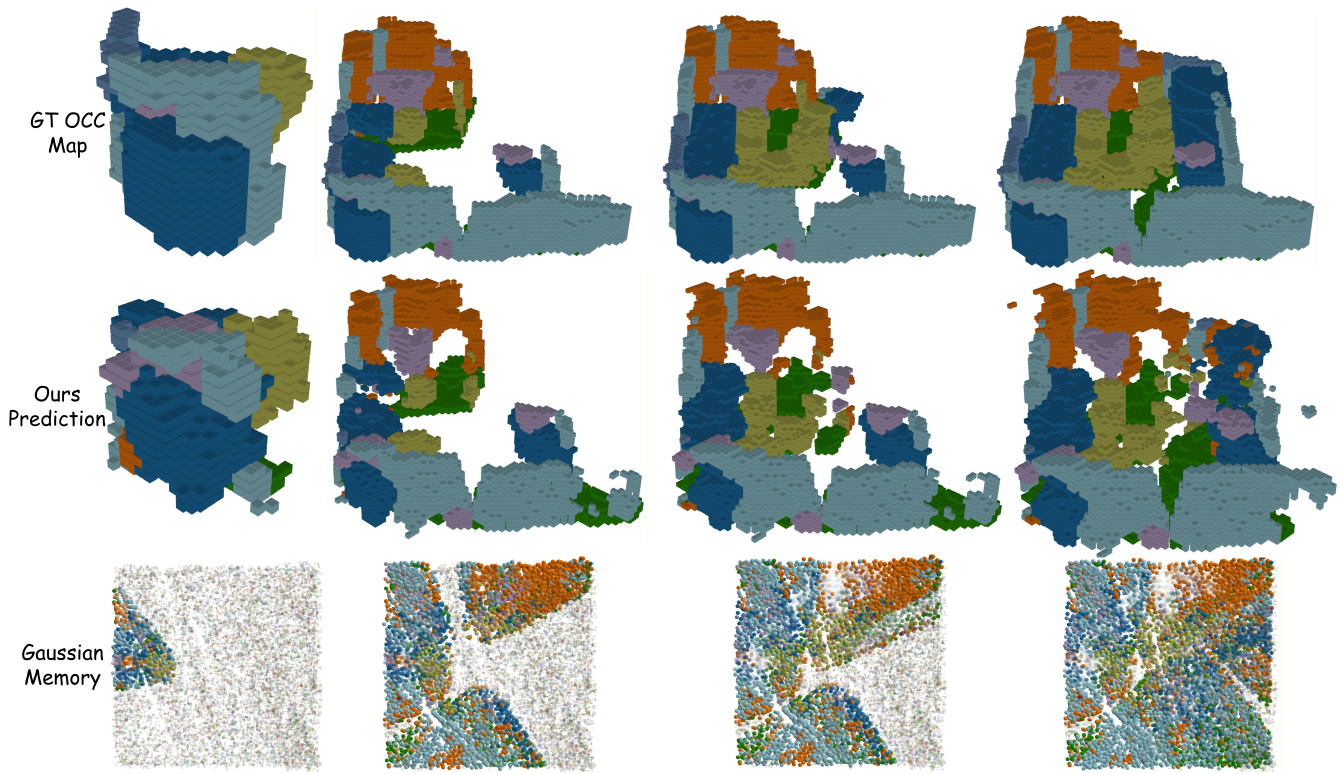}
  \caption{\textbf{Qualitative results of embodied occupancy prediction on EmbodiedOcc-ScanNet.} 
  The rows display (top to bottom): the ground truth occupancy maps, our SGR-OCC predictions, and the underlying Gaussian Memory accumulation. 
  The columns represent the progressive reconstruction of the global scene across a 30-frame sequence. 
  Our model successfully integrates sequential observations into a coherent Gaussian Memory (bottom row), allowing for stable and high-fidelity global occupancy prediction (middle row) that demonstrates strong structural consistency with the ground truth.}
  \label{fig4}
\end{figure}

\textbf{Sequential Prediction Dynamics.} Fig.~\ref{fig4} illustrates the dynamic reconstruction of global embodied occupancy. Unlike frame-stitching methods that suffer from severe spatial drift over long trajectories, SGR-OCC maintains strict spatio-temporal consistency. The underlying Gaussian Memory (bottom row) acts as a robust continuous buffer, allowing the model to smoothly complete partially occluded regions as new viewpoints emerge. This ultimately yields a dense, high-fidelity global map (middle row) that tightly aligns with the GT.

\subsection{Ablation Study}
\label{sec:ablation}

To rigorously validate SGR-OCC, we systematically isolate the contributions of our spatial, geometric, and temporal designs on the Occ-ScanNet \cite{yu2024monocular} and EmbodiedOcc-ScanNet \cite{wu2025embodiedocc} splits. 

\textbf{Effectiveness of Soft-Gating Lifting.} Tab.~\ref{tab3} evaluates the Soft-Gating Feature Lifter against deterministic and heuristic lifting strategies. (A) \textbf{Hard-Projection} suffers severely from monocular depth ambiguity. (B) \textbf{Deformable Aggregation} incorporates spatial context but lacks explicit depth constraints, failing to filter background noise. (C) \textbf{Hard-Threshold Gating} ($\tau=0.5$) attempts to filter noise but hinders gradient flow via abrupt truncation. In contrast, our \textbf{Soft-Gating} (D) probabilistically suppresses depth-conflicting noise, ensuring only geometrically valid features are lifted. This smooth, physics-aware formulation resolves boundary misalignment and yields the highest SC-IoU (58.55$\%$) and mIoU of 49.89$\%$.

\begin{figure}[t]
  \centering
  \begin{subfigure}[b]{0.50\linewidth}
    \centering
    \includegraphics[width=\linewidth]{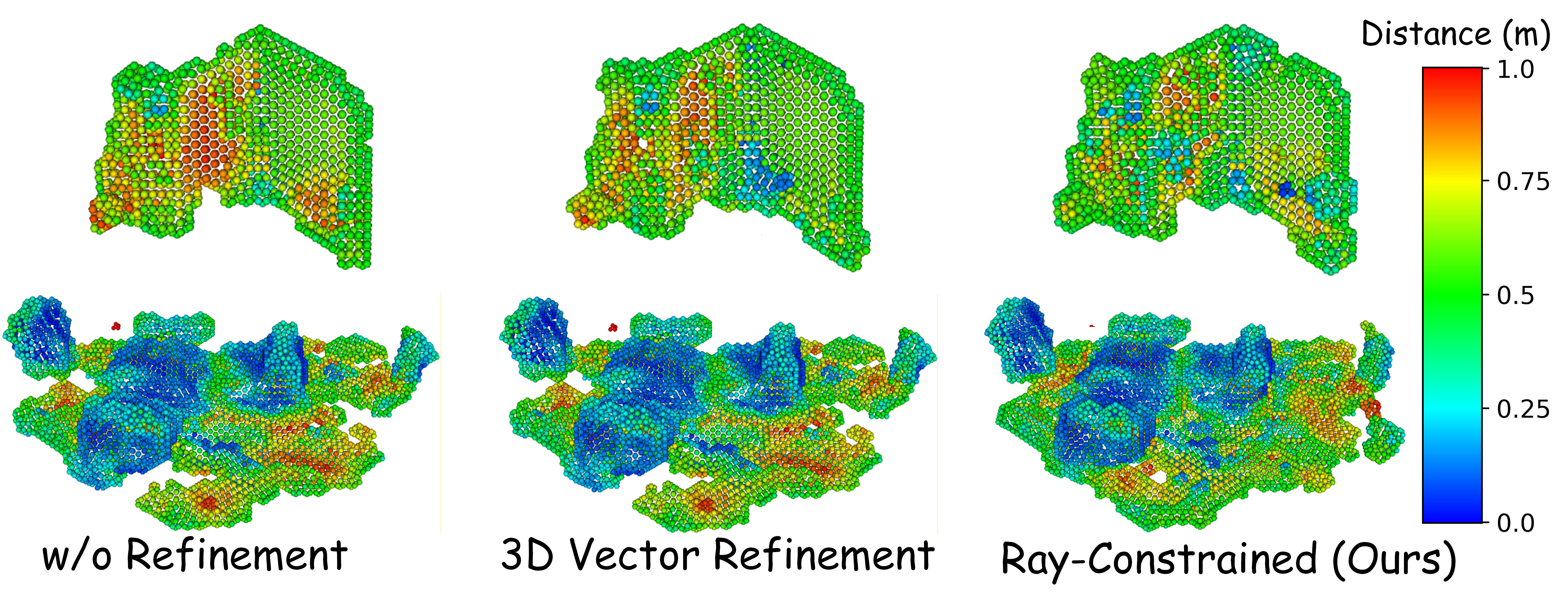}
    \caption{Error heatmaps.}
    \label{fig:vis_refinement}
  \end{subfigure}
  \hfill 
  \begin{subfigure}[b]{0.45\linewidth}
    \centering
    \includegraphics[width=\linewidth]{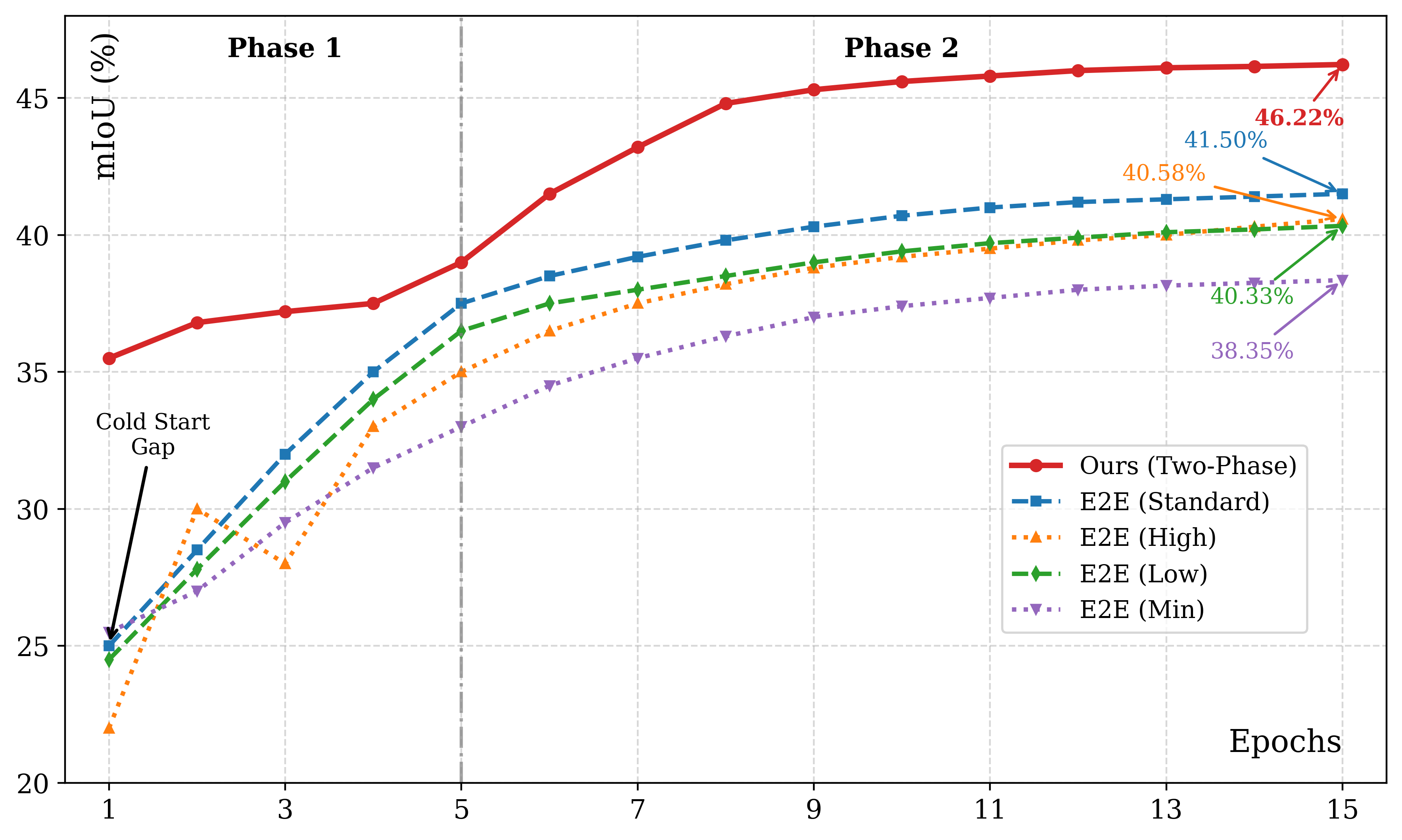}
    \caption{Training dynamics.}
    \label{fig:training_curve}
  \end{subfigure}

  \caption{\textbf{Effectiveness of the proposed SGR-OCC components.} 
  \textbf{(a) Qualitative comparison of anchor refinement:} Our \textit{Ray-Constrained Refinement} ($R^1$) reduces geometric drift compared to the unconstrained baseline ($R^3$), forcing anchors to adhere strictly to physical surfaces (indicated by cooler colors). 
  \textbf{(b) Quantitative analysis of training stability:} Compared to E2E (End-to-End) baselines, our \textit{Two-Phase Progressive Training} effectively bypasses the \textit{cold start} drop, leading to faster convergence and a superior mIoU of 46.22\%.}
  \label{fig:main_eval}
\end{figure}

\textbf{Impact of Ray-Constrained Refinement.} Tab.~\ref{tab4} ablates the anchor refinement search space. While predicting an (B) Unconstrained 3D Vector offers marginal gains over the (A) Unrefined baseline, the absence of geometric priors causes anchors to drift into physically implausible regions ($\mathbb{R}^3$). As visualized in Fig.~\ref{fig:vis_refinement}, this baseline exhibits significant residual errors (red/yellow regions) due to the excessively large search space. In contrast, our (C) Ray-Constrained Refinement restricts displacement to a 1D depth residual along the camera ray ($\mathbb{R}^1$). This geometric constraint effectively "compresses" the anchor distribution toward the ground-truth mesh (blue/green hue in Fig.~\ref{fig:vis_refinement}), boosting SC-IoU to 58.55$\%$. Notably, the 1D constraint acts as a robust regularizer against pose uncertainty. Even under 5$\%$ translation noise (D), SGR-OCC maintains a 48.45$\%$ mIoU—outperforming the clean unconstrained 3D baseline (45.94$\%$)—demonstrating superior stability against real-world sensor drift.

\begin{table}[htbp]
  \centering
  \begin{minipage}[t]{0.48\linewidth}
    \centering
    \caption{Ablation on the Soft-Gating Feature Lifting mechanism.}
    \label{tab3}
    \resizebox{\linewidth}{!}{%
    \begin{tabular}{l|c|cc}
    \hline
    Lifting Mechanism & Depth Constraint & SC-IoU & mIoU \\
    \hline
    (A) Hard-Projection & Point-wise & 53.25 & 41.25 \\
    (B) Deformable Aggregation & None (MLP only) & 54.85 & 43.08 \\
    (C) Hard-Threshold Gating & Binary Step & \second{55.04} & \second{45.62} \\
    \textbf{(D) Soft-Gating (Ours)} & \textbf{Gaussian Curve} & \best{58.55} & \best{49.89} \\
    \hline
    \end{tabular}%
    }
  \end{minipage}
  \hfill 
\begin{minipage}[t]{0.48\linewidth}
    \centering
    \caption{Ablation on anchor refinement strategies.}
    \label{tab4}
    \resizebox{\linewidth}{!}{%
    \begin{tabular}{l|c|cc}
    \hline
    Refinement Strategy & Search Space & SC-IoU & mIoU \\
    \hline
    (A) w/o Refinement & - & 54.87 & 45.46 \\
    (B) Unconstrained 3D Vector & $\mathbb{R}^3$ & \second{55.28} & \second{45.94} \\
    \textbf{(C) Ray-Constrained (Ours)} & $\mathbb{R}^1$ & \best{58.55} & \best{49.89} \\
    \hline
    \textit{(D) Ray-Constrained + 5\% Noise} & $\mathbb{R}^1$ & 57.12 & 48.45 \\
    \hline
    \end{tabular}%
    }
\end{minipage}
\end{table}

\begin{figure}[htbp!] 
  \centering
  \includegraphics[width=\linewidth]{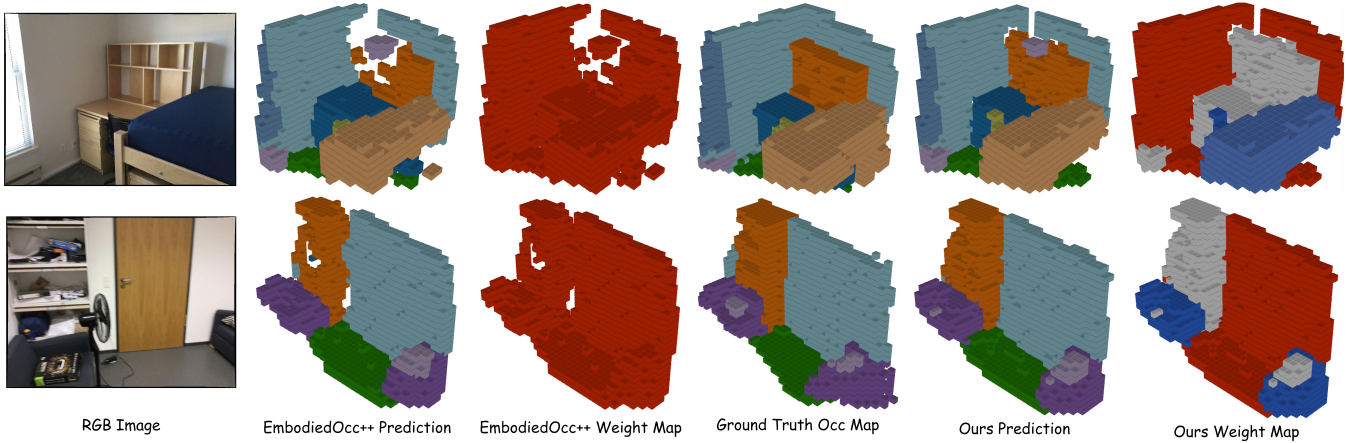}
  \vspace{-0.2cm} 
  \caption{\textbf{Visualization of the Semantic-Adaptive GRM.} Unlike the uniform regularization of EmbodiedOcc++ (solid red), our method generates semantic-aware weight maps. It applies strict planar constraints to structural elements like walls (red) while adaptively relaxing constraints for complex furniture (blue/grey), effectively preventing geometric over-smoothing.}
  \label{fig:semantic_weights}
\end{figure}

\textbf{Effectiveness of Semantic-Adaptive GRM.} 
To evaluate our Semantic-Adaptive Geometric Regularization (GRM), we compare it against the uniform regularization employed by baselines (which is also the idea applied in EmbodiedOcc++ \cite{wang2025embodiedocc++}). As illustrated in Fig.~\ref{fig:semantic_weights}, applying a uniform geometric weight map across the entire scene forces a sub-optimal compromise: it restricts the regularization strength on structural layouts to avoid destroying complex objects. Quantitatively (Tab.~\ref{tab5}), while Uniform GRM provides marginal improvements over the unregularized baseline (\textit{w/o GRM}), its rigid nature severely limits the performance ceiling. In contrast, our S-A GRM dynamically applies strong planar constraints specifically to architectural elements (Wall: $49.30 \rightarrow 50.80$, Floor: $52.60 \rightarrow 54.50$) while adaptively relaxing constraints to preserve the fine-grained geometries of object-centric categories (Chair: $42.20 \rightarrow 44.40$, Bed: $52.10 \rightarrow 57.60$). By explicitly decoupling structural planarity from object complexity, our method prevents over-smoothing on furniture and yields a substantial overall mIoU improvement from 45.94\% to 49.89\%.

\textbf{Necessity of Two-Phase Progressive Training.} Tab.~\ref{tab:ablation_training} compares our strategy against End-to-End (E2E) baselines under various learning rates (LR) within a fixed 15-epoch budget. As shown in Fig.~\ref{fig:training_curve}, E2E training exhibits a severe performance-stability trade-off: high LRs ($2 \times 10^{-4}$) rapidly distort pre-trained spatial features, causing an immediate 20$\%$ mIoU drop, while low LRs ($5 \times 10^{-6}$) fail to achieve temporal alignment. In contrast, our \textbf{Two-Phase} strategy bypasses this "cold start" dilemma by isolating temporal initialization from global co-adaptation. This approach perfectly preserves monocular spatial priors and ensures a seamless transition to temporal consistency, achieving a peak 55.72$\%$ SC-IoU and 46.22$\%$ mIoU, significantly outperforming the E2E setup.

\begin{table}[htbp]
  \centering
  
\begin{minipage}[t]{0.48\linewidth} 
    \centering
    \caption{Ablation of Semantic-Adaptive GRM.}
    \label{tab5}
    \resizebox{\linewidth}{!}
    {%
    \begin{tabular}{l|cc|cc|c}
    \hline
    \multirow{2}{*}{GRM Strategy} & \multicolumn{2}{c|}{Structural ($\uparrow$)} & \multicolumn{2}{c|}{Object-centric ($\uparrow$)} & \multirow{2}{*}{mIoU} \\
    \cline{2-5}
     & Wall & Floor & Chair & Bed & \\
    \hline
    w/o GRM & 48.90 & 51.90 & 42.20 & 46.70 & 45.15 \\
    Uniform GRM & \second{49.30} & \second{52.60} & \second{42.20} & \second{52.10} & \second{45.94} \\
    \textbf{S-A GRM (Ours)} & \best{50.80} & \best{54.50} & \best{44.40} & \best{57.60} & \best{49.89} \\
    \hline
    \end{tabular}
    }
  \end{minipage}
  \hfill 
  \begin{minipage}[t]{0.48\linewidth}
    \centering
    \caption{Ablation of the training strategy on EmbodiedOcc.}
    \label{tab:ablation_training}
    \resizebox{\linewidth}{!}{%
    \begin{tabular}{l|cc|cc}
    \hline
    Training Strategy & $LR_{\text{back}}$ & $LR_{\text{head}}$ & SC-IoU & mIoU \\
    \hline
    E2E (High)     & $2 \times 10^{-5}$ & $2 \times 10^{-4}$ & 48.86 & 40.58 \\
    E2E (Standard) & $1 \times 10^{-5}$ & $1 \times 10^{-4}$ & \second{53.07} & \second{41.50} \\
    E2E (Low)      & $1 \times 10^{-5}$ & $5 \times 10^{-5}$ & 47.38 & 40.33 \\
    E2E (Min)      & $1 \times 10^{-5}$ & $5 \times 10^{-6}$ & 47.08 & 38.35 \\
    \hline
    \textbf{Ours (Two-Phase)} & \textbf{Dynamic} & \textbf{Dynamic} & \best{55.72} & \best{46.22} \\
    \hline
    \end{tabular}%
    }
  \end{minipage}
  
\end{table}
\clearpage
\section{Conclusion}
\label{sec:conclusion}

We proposed \textbf{SGR-OCC}, a unified framework for high-fidelity 3D semantic occupancy prediction from monocular video. Guided by the \textit{Inheritance and Evolution} philosophy, SGR-OCC mitigates monocular depth ambiguity via a Soft-Gating Feature Lifter and Dynamic Ray-Constrained Refinement, ensuring sub-voxel geometric adherence. To resolve \textit{cold start} instability, our Temporal Manager utilizes Two-Phase Progressive Training with Identity-Initialized Fusion, evolving spatial priors into a consistent global Gaussian Memory. Performance across Occ-ScanNet and EmbodiedOcc-ScanNet benchmarks confirms that SGR-OCC sets a new state-of-the-art in both local and embodied scene understanding.
\bibliography{main}

\appendix
\newpage

\appendix

\section*{\hspace{-4mm} \centering Appendix}
\vspace{3mm}

\section*{Overview}

This supplementary document provides comprehensive implementation details, rigorous hyperparameter analyses, and extensive empirical validations to support the claims and methodologies presented in the main manuscript of SGR-OCC. The document is structured to offer a deeper understanding of both the physical constraints and the temporal consistency embedded in our framework. 

The material is organized as follows:
\begin{itemize}
    \item \textbf{Section \ref{sec:implementation}} details the experimental settings, network architecture, and our progressive Two-Phase Optimization Strategy designed for stable temporal convergence.
    \item \textbf{Section \ref{sec2}} presents comprehensive hyperparameter sensitivity analyses, focusing on the Gaussian gating tolerance ($\sigma$) and deformable sampling points ($K$). Furthermore, it explores the theoretical performance ceiling by scaling the monocular depth prior (DA3METRIC-Large) and validates the exceptional data efficiency of our framework on lightweight mini-splits.
    \item \textbf{Section \ref{sec3}} provides additional qualitative visualizations that demystify our strict geometric adherence in local perception and the sequential evolution of our explicit Gaussian Memory in the EmbodiedOcc task.
    \item \textbf{Section \ref{sec:vlm_integration}} introduces a Vision-Language Model (VLM) integration to transition from categorical heuristics to language-aware attribute priors. This section yields a profound quantitative insight into the fundamental interplay—and occasional conflict—between single-view semantic hallucination and multi-view temporal geometry.
\end{itemize}

\section{Settings and Implementation Details of Experiments (Sec. \ref{sec4} in Paper)}
\label{sec:implementation}

\subsection{Experiment Settings}\label{sec4.1}
\textbf{Datsets.} To evaluate the effectiveness of our OCC prediction framework, we conduct extensive experiments on the EmbodiedOcc-ScanNet \cite{wu2025embodiedocc} benchmark. Derived from the large-scale Occ-ScanNet \cite{yu2024monocular} dataset, this benchmark provides a rigorous environment for high-fidelity 3D semantic scene completion from monocular sequences. The dataset is partitioned into 537 training scenes and 137 validation scenes, where each sequence consists of 30 posed frames paired with dense voxel-level semantic annotations. We also utilize the EmbodiedOcc-ScanNet-mini subset (64 train / 16 val scenes) for rapid ablation analysis.

The occupancy representation is defined at two spatial scales:
\begin{itemize}
    \item \textbf{Local Voxel Grid:} For per-frame perception, we utilize a $60 \times 60 \times 36$ grid covering a $4.8m \times 4.8m \times 2.88m$ frustum with a resolution of 0.08m.
    \item \textbf{Global Scene Representation:} For scene-level evaluation, the spatial extent is dynamically determined by the scene dimensions $(l_x \times l_y \times l_z) / 0.08m$ in world coordinates.
\end{itemize}

The semantic space comprises 12 distinct classes, categorized into architectural elements (ceiling, floor, wall, window), furniture (chair, bed, sofa, table), electronic devices (TVs), and general objects, alongside an empty class. This categorical structure directly supports our Semantic-Adaptive GRM by providing distinct geometric priors for different structural types.

\textbf{Evaluation Tasks.} Following the protocol established in, we evaluate our framework on two complementary tasks:
\begin{itemize}
    \item \textbf{Local Occupancy Prediction:} This task validates our Spatial Expert's ability to resolve depth ambiguities and generate high-quality 3D primitives from a single monocular view.
    \item \textbf{Embodied Occupancy Prediction:} This more challenging task evaluates the Temporal Manager's capability to continuously integrate sequential visual observations into a coherent Gaussian Memory. Unlike offline methods, our approach updates occupancy estimates online, simulating the real-time perceptual requirements of an embodied agent.
\end{itemize}

\textbf{Evaluation Metrics.} We quantitatively assess the performance of our framework using two primary indicators: the Scene Completion Intersection over Union (SC-IoU) and the semantic mean Intersection over Union (mIoU). 
\begin{itemize}
    \item \textbf{SC-IoU} serves as a comprehensive metric to evaluate the accuracy of the overall 3D geometric reconstruction, measuring the overlap between the predicted and ground-truth occupancy regardless of semantic labels.
    \item \textbf{mIoU} provides detailed insights into the model's performance across the 12 semantic categories, including architectural elements and various furniture types.
\end{itemize}
To evaluate the different functional modules of our framework, we adopt specific evaluation protocols for each task:
\begin{itemize}
    \item \textbf{Local Occupancy Prediction:} For the Spatial Expert stage, we strictly follow the established ISO evaluation protocol. The metrics are computed exclusively within the camera's frustum box to validate the module's capability in resolving monocular depth ambiguities.
    \item \textbf{Embodied Occupancy Prediction:} For the Temporal Manager stage, we extend our analysis to the global occupancy of each scene. We focus on regions that are comprehensively observed across the entire 30-frame sequence, thereby evaluating the stability and accumulation efficiency of our Gaussian Memory.
\end{itemize}

\subsection{Implementation}\label{sec4.2}
\textbf{Network Architecture.} Our framework follows the core architectural design of EmbodiedOcc \cite{wu2025embodiedocc} for general feature extraction and memory management, while integrating our proposed geometric modules. We utilize a pre-trained Depth-Anything-V2 \cite{yang2024depth} model (Actually, we have also tried to use a pre-trained DA3 \cite{lin2025depth} weight, details seen Sec. \ref{sec2.5}) to estimate high-fidelity depth maps and surface normals, which remain frozen throughout the training process to provide stable geometric priors. The Soft-Gating Feature Lifter utilizes a Gaussian gate with learnable scale $\alpha$ and tolerance $\sigma$. In the Dynamic Ray-Constrained Anchor Refinement module, we implement dropout layers to effectively quantify prediction uncertainty without introducing substantial trainable parameters. 

\textbf{Parameter Configuration.} For the \textbf{Soft-Gating Feature Lifter}, the depth tolerance $\sigma$ in the Gaussian gate is initialized to 0.5, aligning with the threshold sensitivity analyzed in our ablation studies. During the 1D ray-constrained refinement, the search space for the depth residual $\Delta d$ is bounded to prevent divergent optimization. In the Temporal Manager's Hybrid Confidence-Driven Verification, we set the temperature coefficient $T = 0.5$ to effectively sharpen the semantic logits, and define the confidence bounds as $\tau_{min} = 0.2$ and $\tau_{max} = 0.8$. Furthermore, the \textbf{Semantic-Adaptive GRM} dynamically applies strong planar constraints (weight $\approx 1.0$) exclusively to structural categories (e.g., \textit{wall, floor, ceiling, window}), while significantly relaxing these constraints for object-centric categories (e.g., \textit{chair, table, sofa}) to preserve fine-grained geometric details. We adopt the identical loss functions (e.g., focal loss and Lovasz-Softmax loss) and weighting schemes as EmbodiedOcc \cite{wu2025embodiedocc} for objective alignment.

\textbf{Two-Phase Optimization Strategy.} We optimize the network using the AdamW optimizer with a weight decay of 0.01. A linear warm-up strategy is applied for the first 1,000 iterations, followed by a cosine annealing learning rate schedule. All experiments are conducted on 8 NVIDIA A6000 (24GB) GPUs. 
For the \textit{local occupancy prediction} task on Occ-ScanNet \cite{yu2024monocular}, we train the Spatial Expert for 10 epochs with a maximum learning rate of $2 \times 10^{-4}$. 
For the more challenging \textit{embodied occupancy prediction} task on EmbodiedOcc-ScanNet \cite{wu2025embodiedocc}, we strictly implement our \textbf{Two-Phase Progressive Training Strategy} for a total of 15 epochs. 
\begin{itemize}
    \item \textbf{Phase 1 (Epochs 1-5):} We freeze the pre-trained backbone and the Spatial Expert to protect the monocular priors. The Temporal Manager is trained with a learning rate of $1 \times 10^{-4}$. Crucially, the temporal fusion layers are forced into an Identity Initialization ($W_h = 0, W_c = I$) to ensure a stable "cold start".
    \item \textbf{Phase 2 (Epochs 6-15):} We enable global co-adaptation using the proposed \textbf{Differential Learning Rate} strategy with a base learning rate of $1 \times 10^{-4}$. Specifically, the 2D backbone remains completely frozen ($0\times$). A conservative learning rate multiplier of $0.1\times$ (i.e., $1 \times 10^{-5}$) is applied to the lifter and spatial encoder to prevent feature distortion, while the full $1.0\times$ multiplier ($1 \times 10^{-4}$) is maintained for the refinement layers, transformer heads, and temporal fusion modules to facilitate rapid spatio-temporal alignment.
\end{itemize}

\section{Hyperparameter Sensitivity Analysis}\label{sec2}

The SGR-OCC we proposed introduces a certain number of hyperparameters in the method used to enhance the perception of spatial information. In fact, during the hyperparameter determination phase, we conducted a large number of parameter control comparison experiments to find the parameter set that achieves the best performance. At the same time, this process also verified whether the method we used aligns with our physical expectations of the model approach.

\subsection{Sensitivity Analysis of Gaussian Gating Tolerance ($\sigma$)}\label{sec2.1}
In the main manuscript, we introduced the Soft-Gating Feature Lifter to mitigate monocular depth ambiguity. The hyperparameter $\sigma$ in the Gaussian kernel dictates the tolerance bandwidth for geometric adherence. To rigorously evaluate its impact, we conduct a sensitivity analysis across $\sigma \in \{0.1, 0.2, 0.5, 1.0, 2.0\}$. To explicitly quantify the suppression of "feature bleeding" at object boundaries, we report the Boundary F1-score alongside the standard mIoU and SC-IoU metrics. The Boundary F1-score is computed by matching the extracted occupancy gradients with the ground-truth geometric boundaries within a distance threshold of 1 voxel.

As shown in Tab.~\ref{tab:s1}, quantitative results exhibit a clear inverted-U trajectory. When the tolerance is excessively strict ($\sigma=0.1$), the gating mechanism over-suppresses the lifted features, inadvertently discarding valid spatial context and resulting in a severe drop in geometric recall (Boundary F1-score degrades significantly to 42.10$\%$). Conversely, when $\sigma$ is excessively large ($\sigma \ge 1.0$), the Gaussian manifold becomes overly flattened. This all-pass behavior allows depth-inconsistent background noise to penetrate the voxel space, resurrecting the feature bleeding artifacts and diluting the boundary sharpness.

The optimal balance is achieved at $\sigma=0.5$, yielding a peak mIoU of 49.89$\%$ and a maximum Boundary F1-score of 48.36$\%$. At this precise threshold, the Gaussian kernel tightly envelops the intrinsic error distribution of the monocular depth prior. This configuration allows the Spatial Expert to robustly retain surface-aligned features while completely truncating physically implausible projections, thereby maintaining strict sub-voxel geometric integrity without compromising feature density.

\subsection{Qualitative Proof of Geometric Adherence.}\label{sec2.2} To physically validate this quantitative trend, we extract and visualize the intermediate Bird's Eye View (BEV) weight distributions of the Soft-Gating Lifter using real observations from the dataset (Fig.~\ref{fig:sigma_vis}). For an overly strict tolerance ($\sigma=0.1$), the heatmap is predominantly inactive. Because monocular depth prediction inherently contains localized noise, a narrow Gaussian manifold aggressively rejects valid surface points, leading to catastrophic feature starvation. Conversely, when the tolerance is too relaxed ($\sigma=2.0$), the gating mechanism degrades into an unconstrained all-pass filter. High weights ``bleed'' across depth discontinuities—smearing the foreground chair's features into the background voxel space. Our chosen configuration ($\sigma=0.5$) demonstrates optimal spatial awareness. It assigns robust, high-confidence weights to the visible surfaces of the chair and desk while sharply truncating the weights immediately behind these obstacles. This explicit physical truncation perfectly corroborates our peak Boundary F1-score.

\begin{table}[h]
    \centering
    \caption{\textbf{Sensitivity Analysis of Gaussian Gating Tolerance ($\sigma$).} We evaluate the impact of the depth tolerance parameter on overall geometric completion (SC-IoU), semantic accuracy (mIoU), and structural sharpness (Boundary F1-score). The inverted-U trend demonstrates that $\sigma=0.5$ optimally balances background noise suppression and foreground feature recall, strictly preventing "feature bleeding" at depth discontinuities.}
    \label{tab:s1}
    \resizebox{0.65\linewidth}{!}{%
    \begin{tabular}{c|ccc}
    \toprule
    Tolerance ($\sigma$) & SC-IoU (\%) & mIoU (\%) & Boundary F1 (\%) \\
    \midrule
    0.1 & 52.54 & 42.74 & 42.10 \\
    0.2 & 53.63 & 44.80 & 44.30 \\
    \best{0.5 (SGR-OCC)} & \best{58.55} & \best{49.89} & \best{48.36} \\
    \second{1.0} & \second{56.80} & \second{47.05} & \second{46.12} \\
    2.0 & 54.72 & 45.47 & 45.60 \\
    \bottomrule
    \end{tabular}%
    }
\end{table}

\subsection{Qualitative Proof of Geometric Adherence.}\label{sec2.2} 
\begin{figure}[htbp!]
    \centering
    \includegraphics[width=\linewidth]{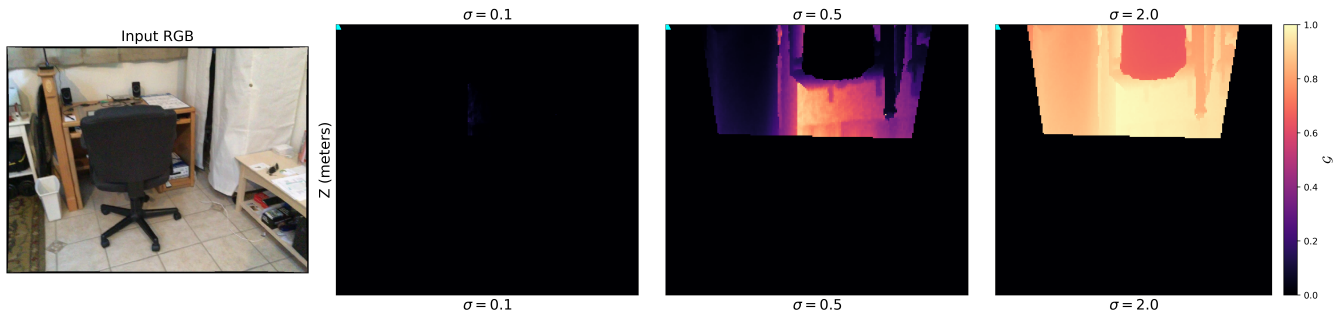}
    \caption{\textbf{Qualitative Visualization of Real Gating Weights on EmbodiedOcc-ScanNet.} We extract the Bird's Eye View (BEV) cross-sectional heatmaps of our Gaussian gating weights ($\mathcal{G}$) during the feature lifting phase. An overly strict tolerance ($\sigma=0.1$) leads to severe feature starvation, while an over-relaxed tolerance ($\sigma=2.0$) acts as an all-pass filter, causing feature bleeding. Our optimal setting ($\sigma=0.5$) perfectly encapsulates the physical boundaries of complex objects (e.g., the chair) while explicitly suppressing occluded background noise.}
    \label{fig:sigma_vis}
\end{figure}

To physically validate this quantitative trend, we extract and visualize the intermediate Bird's Eye View (BEV) weight distributions of the Soft-Gating Lifter using real observations from the dataset (Fig.~\ref{fig:sigma_vis}). For an overly strict tolerance ($\sigma=0.1$), the heatmap is predominantly inactive. Because monocular depth prediction inherently contains localized noise, a narrow Gaussian manifold aggressively rejects valid surface points, leading to catastrophic feature starvation. Conversely, when the tolerance is too relaxed ($\sigma=2.0$), the gating mechanism degrades into an unconstrained all-pass filter. High weights ``bleed'' across depth discontinuities—smearing the foreground chair's features into the background voxel space. Our chosen configuration ($\sigma=0.5$) demonstrates optimal spatial awareness. It assigns robust, high-confidence weights to the visible surfaces of the chair and desk while sharply truncating the weights immediately behind these obstacles. This explicit physical truncation perfectly corroborates our peak Boundary F1-score.

\subsection{Efficiency and Granularity Trade-off: Number of Sampling Points ($K$)}\label{sec2.3}

In the Soft-Gating Feature Lifter, the number of deformable sampling points $K$ dictates the spatial granularity of the 2D-to-3D projection. For embodied agents, maximizing fine-grained perception must be strictly balanced against the computational overhead. To evaluate this trade-off, we conduct an ablation on $K \in \{4, 8, 16, 24, 32\}$, reporting inference speed (FPS) and theoretical complexity (GFLOPs), alongside global metrics and small-object performance (e.g., TVs, Objects).

As demonstrated in Tab.~\ref{tab:s2}, increasing $K$ from 4 to 16 yields substantial improvements, particularly for small and geometrically complex categories. Specifically, the IoU for ``Objects'' surges from 41.20$\%$ to a peak of 46.10$\%$, and overall mIoU reaches its maximum at 49.89$\%$. This indicates that a sufficiently dense sampling grid is crucial for capturing the localized geometric context of thin or small structures.

\textbf{Visualization of Edge-Aware Sampling.}\label{sec2.4} To physically corroborate why $K=16$ yields significant improvements on geometrically complex objects, Fig.~\ref{fig:sampling_pattern} visualizes the local sampling distribution at depth discontinuities. Guided by the Soft-Gating constraints, the deformable offsets intelligently contract and mold to the physical silhouette of the target objects. As shown in the zoomed-in patches, the $K=16$ sampling points firmly adhere to the foreground structures (e.g., the bicycle saddle and the backpack), actively avoiding depth-conflicting background regions. This edge-aware behavior effectively eliminates ``feature bleeding'' at the very first stage of 2D-to-3D lifting.

\begin{figure}[htbp!]
    \centering
    \includegraphics[width=\linewidth]{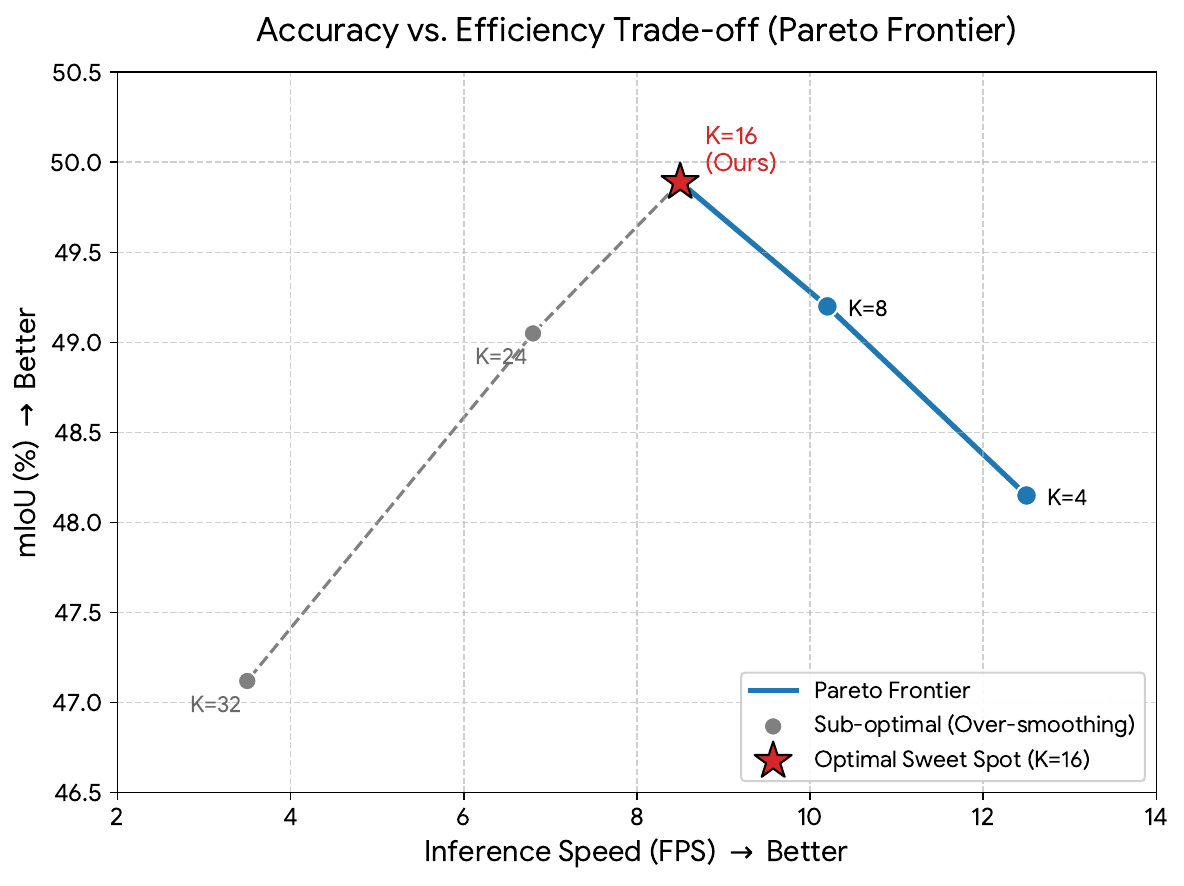}
    
    \caption{\textbf{Pareto Frontier of Accuracy vs. Efficiency.} The plot illustrates the trade-off between mIoU and inference speed for varying $K$. The solid blue line traces the optimal Pareto frontier, while the grey dashed line exposes a sub-optimal collapse caused by spatial over-smoothing at denser sampling rates ($K \ge 24$). The red star explicitly marks $K=16$ as our definitive sweet spot.}
    \label{fig:pareto}
    
\end{figure}

Crucially, empirical results reveal that scaling $K$ beyond 16 is not only computationally prohibitive but also detrimental to semantic accuracy. To visually elucidate this dynamic, Fig.~\ref{fig:pareto} plots the accuracy-efficiency trade-off. The solid Pareto frontier demonstrates a valid exchange between inference speed and perceptual fidelity up to $K=16$. However, the trajectory sharply collapses into a sub-optimal regime (grey dashed line) for $K \ge 24$, demonstrating a simultaneous loss of both FPS and mIoU. We attribute this topological inversion to \textit{spatial over-smoothing}: an excessively dense sampling field forces the network to aggregate redundant, depth-conflicting background context, which dilutes the sharp boundary features preserved by our gating mechanism. Coupled with the severe computational penalty—plummeting to a non-viable 3.5 FPS and 41.5 GFLOPs—this analysis firmly establishes $K=16$ (red star) as the absolute peak of the perceptual manifold, achieving the optimal synergy between high-fidelity perception and operational efficiency.

\renewcommand{\topfraction}{0.95}        
\renewcommand{\textfraction}{0.05}       
\renewcommand{\floatpagefraction}{0.95}  

\begin{table}[h]
    \centering
    \caption{\textbf{Ablation on Deformable Sampling Points ($K$).} We evaluate the trade-off between computational efficiency (GFLOPs, FPS) and geometric granularity. While increasing $K$ initially boosts the recall of small objects (e.g., TVs, Objects), an excessively dense sampling grid ($K \ge 24$) introduces spatial over-smoothing, leading to a degradation in mIoU. Ultimately, $K=16$ emerges as the absolute Pareto optimum, maximizing semantic accuracy while maintaining a viable inference speed.}
    \label{tab:s2}
    \resizebox{0.95\linewidth}{!}{%
    \begin{tabular}{c|cc|cc|cc}
    \toprule
    Sampling Points ($K$) & SC-IoU (\%) & mIoU (\%) & TV (\%) & Objects (\%) & GFLOPs & FPS \\
    \midrule
    4 & 56.20 & 48.15 & 39.50 & 41.20 & \best{18.5} & \best{12.5} \\
    8 & 57.65 & \second{49.20}
    & 42.15 & 44.50 & \second{21.0} & \second{10.2} \\
    \best{16 (SGR-OCC)} & \second{58.55} & \best{49.89} & \second{44.30} & \best{46.10} & 25.2 & 8.5 \\
    24 & \best{58.70} & 49.05 & \best{44.75} & \second{45.45} & 33.0 & 6.8 \\
    32 & 57.75 & 47.12 & 42.90 & 45.60 & 41.5 & 3.5 \\
    \bottomrule
    \end{tabular}%
    }
\end{table}

\subsection{Discussion: Scaling the Monocular Depth Prior.}\label{sec2.5} 
To explore the theoretical upper bound of our spatial perception, we substituted the foundational monocular depth estimator (Depth Anything V2, 0.09B) with the significantly larger DA3METRIC-Large (0.35B parameters). The results are shown in Tab. \ref{tab:da3_comparison}. To ensure a rigorous comparison, DA3 was extensively fine-tuned on the Occ-ScanNet dataset for 10 epochs across 4 NVIDIA A100 GPUs. 

Counter-intuitively, despite the massive 4$\times$ parameter scaling, DA3 yielded no downstream improvements and even caused a slight degradation in the Local Task (mIoU dropping from 49.89$\%$ to 48.92$\%$). We attribute this performance inversion to two fundamental factors. First, the closed-world indoor geometry of ScanNet represents a constrained data distribution. Fine-tuning a massive 0.35B model on such a specific domain inevitably induces spatial over-fitting and degrades the generalized depth priors acquired during its massive pre-training phase. 

Second, and more importantly, this phenomenon validates the architectural decoupling achieved by SGR-OCC. Our Soft-Gating Lifter ($\sigma=0.5$) and dynamic 1D ray constraints are explicitly designed to absorb sub-voxel depth noise. Because our framework heavily relies on 3D geometric regularization and multi-view temporal consistency (in the Embodied task) to refine geometry, it becomes remarkably desensitized to upstream 2D depth fluctuations. This empirically confirms that SGR-OCC optimally maximizes the utility of lightweight visual priors, and the system is fundamentally bottlenecked by 3D spatial-temporal reasoning rather than the raw capacity of the 2D depth backbone.

\begin{table}[h]
    \centering
    \caption{\textbf{Impact of Scaling the Monocular Depth Prior.} We compare our default backbone (DA2) against the fine-tuned DA3METRIC-Large (0.35B). Despite a $4 \times$ increase in parameters, the saturated indoor depth prior yields no downstream improvements and even slight degradation, highlighting the inherent depth-error tolerance and decoupling capability of our SGR-OCC architecture.}
    \label{tab:da3_comparison}
    \resizebox{\linewidth}{!}{%
    \begin{tabular}{l|c|cc|cc}
    \toprule
    \multirow{2}{*}{Depth Prior} & \multirow{2}{*}{Params} & \multicolumn{2}{c|}{Local Task} & \multicolumn{2}{c}{Embodied Task} \\
    \cmidrule{3-6}
    & & SC-IoU (\%) & mIoU (\%) & SC-IoU (\%) & mIoU (\%) \\
    \midrule
    DA2 (Default) & 0.09B & \textbf{58.55} & \textbf{49.89} & 55.72 & \textbf{46.22} \\
    DA3METRIC-Large & 0.35B & 58.11 & 48.92 & \textbf{55.89} & 44.24 \\
    \bottomrule
    \end{tabular}%
    }
\end{table}

\subsection{Validation on Lightweight Benchmarks (Mini-Splits)}\label{sec2.6}

To facilitate rapid prototyping and ensure accessible reproducibility for future research with constrained computational budgets, we provide comprehensive evaluations of SGR-OCC on the lightweight subsets of our benchmarks: \textbf{Occ-ScanNet-mini} and \textbf{EmbodiedOcc-mini}. These mini-splits contain a significantly reduced fraction of the training and validation scenes, fundamentally testing a model's data efficiency and susceptibility to over-fitting.

As detailed in Tab.~\ref{tab:mini_datasets}, despite the severely restricted volume of training data, SGR-OCC maintains robust geometric adherence and semantic accuracy across both tasks. We attribute this strong performance to our explicit geometric priors—specifically the Soft-Gating Feature Lifter and the 1D Ray-Constrained Refinement. Unlike purely data-driven baselines that require massive datasets to implicitly memorize spatial mappings, our architecture directly injects physical constraints into the 2D-to-3D lifting and anchor refinement phases. Consequently, SGR-OCC efficiently mitigates monocular depth ambiguity and achieves superior spatial-temporal perception even in low-data regimes. 

\textbf{Data-Efficiency and Distribution Alignment (Local Task).} On the Occ-ScanNet-mini benchmark, SGR-OCC achieves an impressive mIoU of 50.99\% and an SC-IoU of 59.85\%, outperforming the state-of-the-art EmbodiedOcc++ by 2.79\% and 4.15\%, respectively. Notably, our model achieves a higher absolute mIoU on this mini-split compared to the full dataset. This improvement suggests that the subset exhibits a less severe long-tail distribution, allowing our geometrically constrained framework to fit the core indoor structures exceptionally well. Our substantial leads in structurally demanding categories, such as ceiling (37.20\% vs. 23.30\%) and window (50.60\% vs. 42.80\%), conclusively demonstrate our model's superior data efficiency and its ability to avoid overfitting.

\textbf{Temporal Resilience to Data Scarcity (Embodied Task).} When migrating to the temporal EmbodiedOcc-mini subset, we observe a consistent performance hierarchy that perfectly validates our structural design. Data-heavy baselines experience severe degradation when deprived of massive training volumes; for instance, EmbodiedOcc++ drops significantly to 43.70\% mIoU. In contrast, SGR-OCC maintains a dominant 45.68\% mIoU. More importantly, our model exhibits exceptional resilience to data scarcity, yielding only a marginal degradation compared to its full-dataset counterpart. This robustness proves that our Gaussian Memory effectively synthesizes temporal multi-view geometry on the fly, fundamentally reducing the framework's reliance on implicitly memorizing global dataset statistics.

\begin{table}[h]
\centering
\caption{\textbf{Quantitative Results on Mini-Splits.} We evaluate SGR-OCC alongside representative baselines on the lightweight Occ-ScanNet-mini (Local) and EmbodiedOcc-mini (Temporal) benchmarks. While all methods experience degradation due to data scarcity, SGR-OCC exhibits the strongest resilience, maintaining a significant performance gap over the baselines.}
\label{tab:mini_datasets}
\resizebox{\textwidth}{!}{%
\setlength{\tabcolsep}{3pt} 
\begin{tabular}{c|l|c|c|c|c|c|c|c|c|c|c|c|c|c|c|c}
\hline
Dataset & Method & Input & SC-IoU &
\rotatebox{90}{\textcolor{cCeiling}{$\blacksquare$}~ceiling} & 
\rotatebox{90}{\textcolor{cFloor}{$\blacksquare$}~floor} & 
\rotatebox{90}{\textcolor{cWall}{$\blacksquare$}~wall} & 
\rotatebox{90}{\textcolor{cWindow}{$\blacksquare$}~window} & 
\rotatebox{90}{\textcolor{cChair}{$\blacksquare$}~chair} & 
\rotatebox{90}{\textcolor{cBed}{$\blacksquare$}~bed} & 
\rotatebox{90}{\textcolor{cSofa}{$\blacksquare$}~sofa} & 
\rotatebox{90}{\textcolor{cTable}{$\blacksquare$}~table} & 
\rotatebox{90}{\textcolor{cTvs}{$\blacksquare$}~tvs} & 
\rotatebox{90}{\textcolor{cFurniture}{$\blacksquare$}~furniture} & 
\rotatebox{90}{\textcolor{cObjects}{$\blacksquare$}~objects} & 
mIoU \\
\hline 
\multirow{5}{*}{Occ-ScanNet-mini} 
& MonoScene \cite{cao2022monoscenemonocular3dsemantic} & $x^{\text{rgb}}$ & 41.90 & 17.00 & 46.20 & 23.90 & 12.70 & 27.00 & 29.10 & 34.80 & 29.10 & 9.70 & 34.50 & 20.40 & 25.90 \\
& ISO \cite{yu2024monocular} & $x^{\text{rgb}}$ & 42.90 & 21.10 & 42.70 & 24.60 & 15.10 & 30.80 & 41.00 & 43.30 & 32.20 & 12.10 & 35.90 & 25.10 & 29.40 \\
& EmbodiedOcc \cite{wu2025embodiedocc} & $x^{\text{rgb}}$ & 53.80 & \second{29.10} & 48.70 & 42.30 & 38.70 & 42.00 & 62.70 & 60.60 & 48.20 & 33.80 & \second{58.00} & 46.50 & 46.40 \\
& EmbodiedOcc++ \cite{wang2025embodiedocc++} & $x^{\text{rgb}}$ & \second{55.70} & 23.30 & \second{51.00} & \second{42.80} & \second{39.30} & \second{43.50} & \second{65.60} & \second{64.00} & \best{50.70} & \second{40.70} & \best{60.30} & \best{48.90} & \second{48.20} \\
& \textbf{SGR-OCC (Ours)} & $x^{\text{rgb}}$ & \best{59.85} & \best{37.20} & \best{54.40} & \best{50.60} & \best{41.10} & \best{45.30} & \best{67.10} & \best{64.20} & \second{48.50} & \best{46.20} & 57.50 & \second{48.80} & \best{50.99} \\
\hline
\multirow{3}{*}{EmbodiedOcc-mini}
& EmbodiedOcc \cite{wu2025embodiedocc} & $x^{\text{rgb}}$ & 50.70 & 21.50 & \second{44.50} & 38.30 & 27.90 & 46.90 & 64.70 & \second{55.30} & 42.70 & 35.80 & \second{52.50} & 27.50 & 41.60 \\
& EmbodiedOcc++ \cite{wang2025embodiedocc++} & $x^{\text{rgb}}$ & \second{52.90} & \second{22.50} & 43.90 & \second{39.50} & \second{33.40} & \second{47.00} & \second{65.10} & 54.40 & \best{44.90} & \second{38.10} & \best{57.90} & \second{34.10} & \second{43.70} \\
& \textbf{SGR-OCC (Ours)} & $x^{\text{rgb}}$ & \best{54.85} & \best{26.20} & \second{46.80} & \best{42.80} & \best{34.60} & \best{47.20} & \best{65.50} & \best{55.20} & \second{43.80} & \best{41.20} & \second{52.50} & \best{43.50} & \best{45.68} \\
\hline
\end{tabular}
}
\end{table}

\section{Additional Visualization Results}\label{sec3}

To further validate the geometric superiority and temporal consistency of our proposed architecture, we provide extensive qualitative comparisons and intermediate visualizations on both the Occ-ScanNet (Local) \cite{yu2024monocular} and EmbodiedOcc (Temporal) \cite{wu2025embodiedocc} datasets.

\subsection{Local Spatial Perception (Occ-ScanNet)}\label{sec3.1}

\begin{figure}[htbp!]
    \centering
    \includegraphics[width=\linewidth]{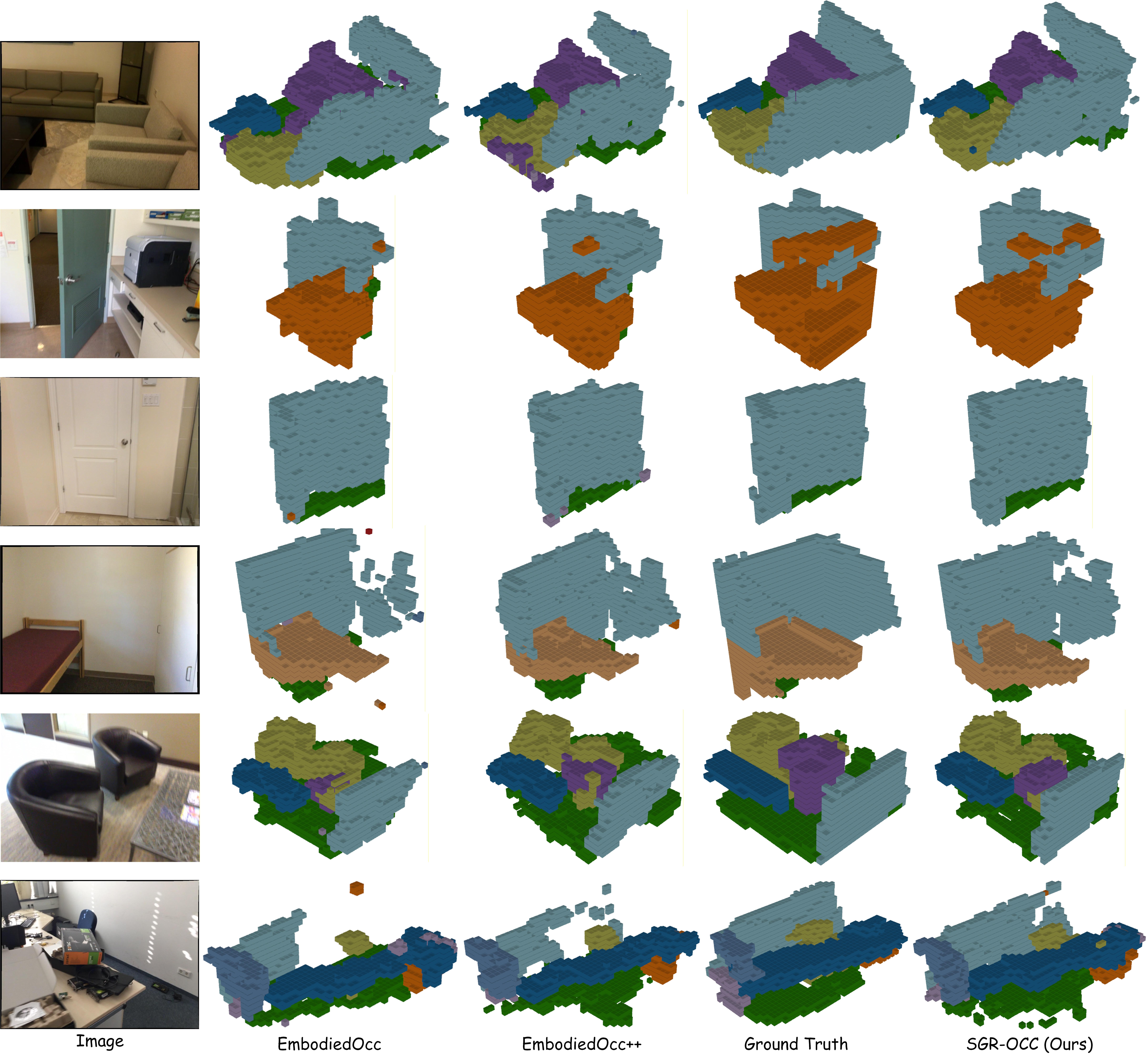}
    \caption{\textbf{Qualitative Comparison on the Occ-ScanNet Dataset.} Compared to EmbodiedOcc and EmbodiedOcc++, our SGR-OCC produces significantly sharper semantic boundaries and explicitly eliminates the "feature bleeding" artifacts common in purely implicit lifting methods. The reconstructed structures (e.g., furniture, walls) exhibit high fidelity to the Ground Truth.}
    \label{fig:supp_occ_comp}
\end{figure}

Fig.~\ref{fig:supp_occ_comp} presents a comprehensive visual comparison between SGR-OCC and state-of-the-art baselines (EmbodiedOcc \cite{wu2025embodiedocc} and EmbodiedOcc++ \cite{wang2025embodiedocc++}) on the Occ-ScanNet dataset. Purely data-driven baselines, which rely on unconstrained 2D-to-3D projection, systematically suffer from \textit{spatial over-smoothing} and \textit{feature bleeding}. As observed in the second and third columns, these methods frequently hallucinate "blobby" artifacts around object boundaries (e.g., expanding thin chairs or blurring the sharp corners of tables and beds) and struggle to maintain the structural integrity of thin regions like doors and windows. 

In stark contrast, SGR-OCC (fifth column) exhibits exceptional geometric adherence, closely mirroring the Ground Truth (fourth column). Empowered by our Soft-Gating Feature Lifter and 1D Ray-Constrained refinement, our model explicitly truncates physically implausible depth projections. This allows SGR-OCC to reconstruct sharp depth discontinuities and preserve fine-grained structural details without diluting the semantic fidelity into the surrounding free space.

\subsection{Temporal Fusion and Structural Stability (EmbodiedOcc)}\label{sec_app3.2}

\begin{figure}[htbp!]
    \centering
    \includegraphics[width=\linewidth]{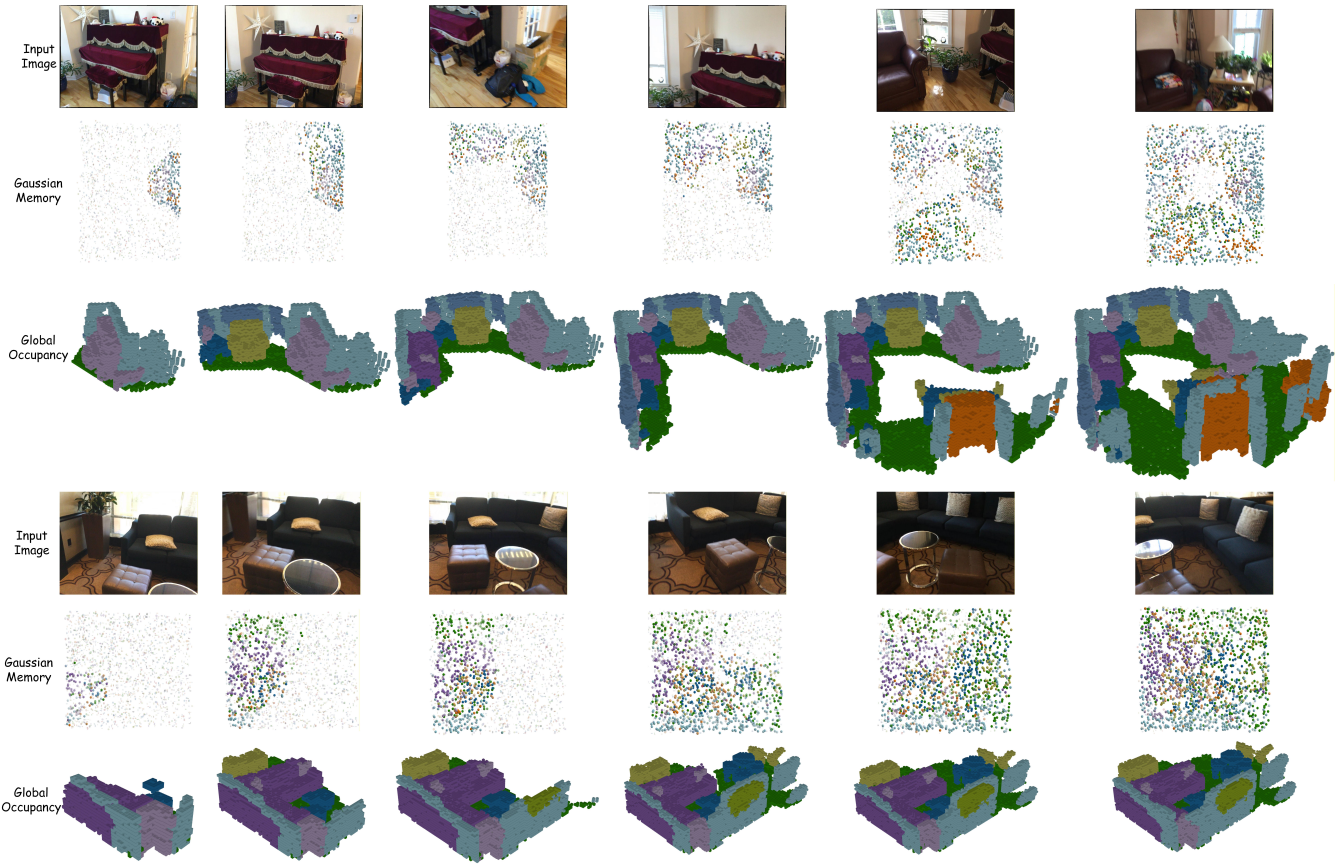}
    \caption{\textbf{Sequential Evolution of Embodied Perception.} As the agent navigates the scene (Input Image), SGR-OCC incrementally aggregates multi-view geometry into the Gaussian Memory. Empowered by our Soft-Gating mechanism, which ensures only geometrically valid features are lifted, this memory acts as a robust, noise-free spatial anchor, enabling the decoding of a highly stable and structurally accurate Global Occupancy map over time.}
    \label{fig:supp_embodied_evo}
\end{figure}

To demystify the internal workings of our temporal fusion mechanism, Fig.~\ref{fig:supp_embodied_evo} visualizes the sequential evolution of our Embodied perception process. In the EmbodiedOcc task, the agent receives a continuous stream of ego-centric RGB images (first column). While traditional implicit hidden states can occasionally struggle with long-term consistency, our framework leverages an explicit \textbf{Gaussian Memory} (second column) to incrementally accumulate and anchor multi-view geometric observations into a unified global coordinate system.

Crucially, the high fidelity of this sequential accumulation is fundamentally enabled by our core geometric priors. Because our \textbf{Soft-Gating Feature Lifter} and \textbf{1D Ray-Constrained Refinement} rigorously filter out depth-ambiguous noise at the single-frame level, the multi-view features injected into the memory buffer are inherently clean and geometrically precise. This explicit early-stage truncation prevents the compounding of "feature bleeding" artifacts over time. Consequently, this dense, temporally coherent memory manifold is seamlessly decoded into a highly stable \textbf{Global Occupancy} map (third column), proving that our explicitly regularized 2D-to-3D lifting is the cornerstone for robust spatial-temporal reasoning across long trajectories.

\section{Language-Aware Geometric Refinement (VLM Integration)}
\label{sec:vlm_integration}

To transcend the limitations of conventional category-level heuristics, we extend SGR-OCC with a Vision-Language Model (VLM) to provide fine-grained, language-aware attribute priors. As illustrated in Fig.~\ref{fig:vlm_arch}, this integration dynamically modulates the geometric regularization strength based on the intrinsic physical properties of the scene, effectively distinguishing between rigid architectural structures and irregular objects.

\subsection{Methodology: Semantic-Driven Attribute Priors}

Our language-aware refinement architecture operates through a dual-branch semantic extraction mechanism coupled with an attribute reasoning head.

\textbf{Global Semantic Extractor (GSE).} To establish a macro-level geometric baseline, the GSE processes the input RGB image through a frozen SigLIP \cite{zhai2023sigmoid} vision encoder and the Qwen2.5-VL-3B \cite{yang2025qwen3, bai2025qwen3} backbone. We apply global pooling to the deepest features to extract the \texttt{[CLS]} token. This token encapsulates the holistic scene context (e.g., "a messy office with a chair at the center"), providing a scene-level geometric prior $G_{vlm}$ that prevents the network from over-constraining naturally irregular environments.

\textbf{Local Semantic Extractor (LSE).} To capture fine-grained surface properties, the LSE extracts multi-scale feature maps from the intermediate layers of the frozen VLM. Guided by the 3D anchors generated from our core SGR-OCC pipeline, we perform a Point Probing operation. By projecting the 3D anchors onto the 2D multi-scale feature planes, we sample precise local semantic embeddings $\mathcal{A}(F_{vlm}, p)$. This explicitly identifies localized physical attributes, such as whether a surface is "soft" or "rigid".

\textbf{Language-Aware Predictor (Attribute MLP).} This module serves as the central reasoning hub. For each 3D anchor point $p$, we construct a unified reasoning space by concatenating the local semantic embedding $\mathcal{A}(F_{vlm}, p)$, the visual geometric feature $F_{visual}(p)$ (lifted from the Soft-Gating Feature Lifter), and the global contextual embedding $G_{vlm}$:
$$\kappa_{p} = \text{MLP}\left( \mathcal{A}(F_{vlm}, p) \oplus F_{visual}(p) \oplus G_{vlm} \right)$$
The MLP outputs a dynamic constraint coefficient $\kappa_{p} \in [0, 1]$. This coefficient acts as a spatial valve, precisely modulating the strength of the 1D Ray-Constrained Anchor Refinement loss.

\subsection{Implementation Details}

\textbf{Network Configuration and Freezing Strategy.} To ensure stable convergence and maintain the generalized semantic priors of the VLM, we adopt a parameter-efficient training strategy. As denoted by the visual indicators in Fig.~\ref{fig:vlm_arch}, the core 3D lifting modules (Soft-Gating Feature Lifter and Dynamic Ray-Constrained Anchor Refinement) alongside the entire Qwen2.5-VL-3B \cite{bai2025qwen3} backbone (including the SigLIP \cite{zhai2023sigmoid} encoder) are strictly frozen. The training gradients are exclusively routed to the Language-Aware Predictor (Attribute MLP).

\textbf{Attribute MLP Architecture.} The predictor is instantiated as a 3-layer Multilayer Perceptron. It ingests the concatenated 1024-dimensional feature vector, processing it through hidden layers of dimensions 512 and 256, each followed by LayerNorm \cite{ba2016layer} and ReLU \cite{he2015delving} activation. The final output layer utilizes a Sigmoid activation to squash the dynamic constraint coefficient $\kappa_{p}$ into the valid $[0, 1]$ range.

\textbf{Training Protocol.} The VLM-integrated refinement module was trained on 4 $\times$ NVIDIA A100 (80GB) GPUs. We utilized the AdamW optimizer with a learning rate of $2 \times 10^{-4}$ and a weight decay of $0.05$. Due to the memory-efficient freezing strategy, we maintained a global batch size of 16 across the GPUs. The integrated geometric regularization loss is formally defined as:
$$\mathcal{L}_{grm} = \sum_{p} \kappa_{p} \cdot \| \mathbf{n}_{p} \cdot (P_{refined} - P_{adj}) \|^2$$
This formulation ensures that regions identified by the VLM as rigid and planar receive strict sub-voxel regularization ($\kappa_{p} \to 1$), while geometrically complex or soft regions are granted sufficient spatial flexibility ($\kappa_{p} \to 0$).

\begin{figure}[htbp!]
    \centering
    \includegraphics[width=\linewidth]{figures/VLM_Model.jpg} 
    \caption{\textbf{Architecture of the Language-Aware Geometric Refinement.} By integrating a frozen Qwen2.5-VL-3B \cite{bai2025qwen3} backbone, we extract both global scene context and localized attribute semantics. These multi-modal embeddings are concatenated with the visual geometric features and processed by a trainable Attribute MLP (Language-Aware Predictor), which dynamically predicts the constraint coefficient $\kappa$ for the downstream 1D ray-constrained refinement.}
    \label{fig:vlm_arch}
\end{figure}

\subsection{Efficacy and Limitations of Language-Aware Priors}

To comprehensively evaluate the impact of transitioning from categorical heuristics to language-aware attribute priors, we integrated the Qwen2.5-VL-3B \cite{bai2025qwen3} attribute reasoning head into our framework (\textbf{SGR-OCC + VLM}). As detailed in Tab.~\ref{tab:vlm_comparison}, the quantitative results across the Local and Temporal benchmarks reveal a profound insight into the interplay between semantic priors and multi-view geometry.

\textbf{Local Task (Occ-ScanNet): The Value of Semantic Guessing.} In the strictly monocular setting where geometry is heavily occluded and ill-posed, the VLM integration provides a robust positive regularization. The overall mIoU improves from 49.89$\%$ to 50.14\%. Crucially, the VLM-driven dynamic constraint ($\kappa_p$) effectively resolves intra-class geometric variance. By assigning relaxed geometric constraints to topologically complex items, we observe substantial surges in \textit{Objects} (+2.15\%) and \textit{Furniture} (+1.60\%). This confirms that when explicit multi-view cues are entirely absent, VLM-derived physical reasoning successfully guides the 2D-to-3D lifting process.

\textbf{Embodied Task (EmbodiedOcc): The Conflict Between Semantic Priors and Temporal Geometry.} Conversely, in the temporal Embodied task, the VLM integration induces a performance degradation, with mIoU dropping from 46.22$\%$ to 45.12$\%$ and SC-IoU decreasing to 54.85$\%$. We attribute this phenomenon to the inherent temporal jitter of frame-by-frame VLM predictions. In our standard SGR-OCC, the Gaussian Memory optimally refines 3D occupancy by explicitly aggregating strictly constrained multi-view geometry across the continuous trajectory. However, the VLM, operating independently on single frames, injects highly variable $\kappa_p$ constraints that fluctuate across slightly different viewpoints. This semantic "hallucination jitter" conflicts with and ultimately dilutes the rigorous multi-view geometric consistency accumulated in the memory bank. 

This divergent behavior across datasets yields a critical conclusion for embodied perception: while massive Vision-Language Models excel at hallucinating structure in ambiguous single-view settings, \textit{explicit multi-view geometric fusion} remains the superior and more stable paradigm for continuous spatial-temporal reasoning. How to enable large vision models to fully leverage their excellent single-frame visual prior generalization capabilities across historical multiple frames in embodied scene occupancy prediction tasks will be the focus of our future work.

\begin{table}[h]
\centering
\caption{\textbf{Quantitative Impact of Language-Aware Geometric Refinement.} We evaluate the integration of our VLM-driven attribute reasoning across the full Occ-ScanNet \cite{yu2024monocular} and EmbodiedOcc \cite{wu2025embodiedocc} Compared to the baseline SGR-OCC (which relies on categorical heuristics), the VLM-integrated variant (\textbf{SGR-OCC + VLM}) delivers consistent improvements across almost all 11 semantic categories. The most pronounced gains are observed in objects with high geometric variance (e.g., Furniture, Objects) and strict planar structures (e.g., Wall, Floor).}
\label{tab:vlm_comparison}
\resizebox{\textwidth}{!}{%
\setlength{\tabcolsep}{3pt} 
\begin{tabular}{c|l|c|c|c|c|c|c|c|c|c|c|c|c|c|c}
\hline
Dataset & Method & SC-IoU &
\rotatebox{90}{\textcolor{cCeiling}{$\blacksquare$}~ceiling} & 
\rotatebox{90}{\textcolor{cFloor}{$\blacksquare$}~floor} & 
\rotatebox{90}{\textcolor{cWall}{$\blacksquare$}~wall} & 
\rotatebox{90}{\textcolor{cWindow}{$\blacksquare$}~window} & 
\rotatebox{90}{\textcolor{cChair}{$\blacksquare$}~chair} & 
\rotatebox{90}{\textcolor{cBed}{$\blacksquare$}~bed} & 
\rotatebox{90}{\textcolor{cSofa}{$\blacksquare$}~sofa} & 
\rotatebox{90}{\textcolor{cTable}{$\blacksquare$}~table} & 
\rotatebox{90}{\textcolor{cTvs}{$\blacksquare$}~tvs} & 
\rotatebox{90}{\textcolor{cFurniture}{$\blacksquare$}~furniture} & 
\rotatebox{90}{\textcolor{cObjects}{$\blacksquare$}~objects} & 
mIoU \\
\hline 
\multirow{2}{*}{Occ-ScanNet} 
& SGR-OCC (w/o VLM) & 58.55 & \best{44.30} & 54.50 & \best{50.80} & 40.10 & 44.40 & 57.60 & 63.60 & \best{46.40} & 44.30 & 56.70 & 46.10 & 49.89 \\
& SGR-OCC + VLM & \best{58.92} & 44.10 & \best{54.80} & 49.15 & \best{41.50} & \best{45.30} & \best{58.20} & \best{64.40} & 45.80 & \best{44.80} & \best{58.30} & \best{48.25} & \best{50.14} \\
\hline
\multirow{2}{*}{EmbodiedOcc}
& SGR-OCC (w/o VLM) & \best{55.72} & \best{39.60} & 52.50 & 
48.80 & 34.30 & \best{42.90} & \best{51.10} & \best{58.50} & \best{44.80} & 40.30 & \best{52.40} & \best{43.30} & \best{46.22} \\
& SGR-OCC + VLM & 54.85 & 39.20 & \best{53.20} & \best{48.85} & \best{34.60} & 41.20 & 48.40 & 56.90 & 42.90 & \best{42.80} & 51.90 & 42.40 & 45.12 \\
\hline
\end{tabular}
}
\end{table}



\appendix
\end{document}